\PassOptionsToPackage{table}{xcolor}

\documentclass[10pt,twocolumn,letterpaper]{article}

\usepackage[pagenumbers]{wacv} %

\usepackage{graphicx}
\usepackage{amsmath}
\usepackage{amssymb}
\usepackage{booktabs}
\usepackage{comment}

\usepackage{times}
\usepackage{epsfig}
\usepackage{float}
\usepackage{multirow}
\usepackage{amsfonts}
\usepackage{graphics}
\usepackage{witharrows}
\usepackage{arydshln}
\usepackage{amsfonts} 
\usepackage{mathtools}
\usepackage{enumitem}

\usepackage[accsupp]{axessibility}  %

\usepackage[pagebackref,breaklinks,colorlinks]{hyperref}

\usepackage[capitalize]{cleveref}
\crefname{section}{Sec.}{Secs.}
\Crefname{section}{Section}{Sections}
\Crefname{table}{Table}{Tables}
\crefname{table}{Tab.}{Tabs.}

\newcommand{\fref}[1]{Figure \ref{#1}}
\newcommand{\tref}[1]{Table \ref{#1}}

\newcolumntype{g}{>{\columncolor[HTML]{EFEFEF}}c}
\newcolumntype{q}{>{\columncolor[HTML]{EFEFEF}}l}

\newcommand*\samethanks[1][\value{footnote}]{\footnotemark[#1]}

\def\wacvPaperID{1329} %
\def\confName{WACV}
\def\confYear{2025}

\begin{document}

\title{{\color{red}Mi}xed {\color{red}Pa}tch Visible-Infrared Modality Agnostic Object Detection}

\author{Heitor Rapela Medeiros\thanks{Equal contribution. Email: \{heitor.rapela-medeiros.1, david.latortue\}@ens.etsmtl.ca}, David Latortue\samethanks, Eric Granger, Marco Pedersoli \\ LIVIA, Dept. of Systems Engineering, ETS Montreal, Canada}

\maketitle

\begin{abstract}
In real-world scenarios, using multiple modalities like visible (RGB) and infrared (IR) can greatly improve the performance of a predictive task such as object detection (OD). Multimodal learning is a common way to leverage these modalities, where multiple modality-specific encoders and a fusion module are used to improve performance. In this paper, we tackle a different way to employ RGB and IR modalities, where only one modality or the other is observed by a single shared vision encoder. This realistic setting requires a lower memory footprint and is more suitable for applications such as autonomous driving and surveillance, which commonly rely on RGB and IR data. However, when learning a single encoder on multiple modalities, one modality can dominate the other, producing uneven recognition results. This work investigates how to efficiently leverage RGB and IR modalities to train a common transformer-based OD vision encoder, while countering the effects of modality imbalance. For this, we introduce a novel training technique to Mix Patches (MiPa) from the two modalities, in conjunction with a patch-wise modality agnostic module, for learning a common representation of both modalities. Our experiments show that MiPa can learn a representation to reach competitive results on traditional RGB/IR benchmarks while only requiring a single modality during inference. Our code is available at: \url{https://github.com/heitorrapela/MiPa}.
\end{abstract}

\begin{figure*}[!ht]
\centering
    \resizebox{\textwidth}{!}{%

    \begin{tabular}{c}
    
    \includegraphics[width=1.0\textwidth]{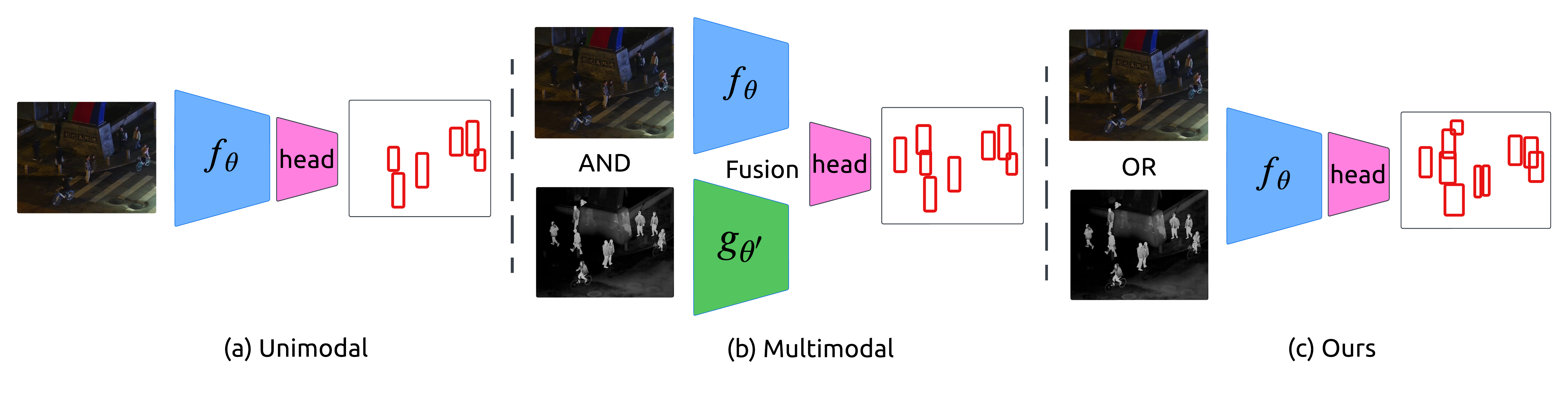} \\
    
    \end{tabular}
    }
\caption{Differences in inputs for different modality learning. (a) \textit{Unimodal} learning assumes that only one modality is used for both training and testing. (b) \textit{Multimodal} learning requires multiple modalities and a special architecture to fuse them in order to improve performance. (c) \textit{Ours} assumes that a model should be able to perform well for both modalities by using both for training but only one at a time for testing and with a shared vision encoder.}
\label{fig:cpirgb_different_learning}
\end{figure*}

\section{Introduction} 
\label{sec:intro}

\newcommand{\cmmnt}[1]{}

In recent years, the reducing costs in data acquisition and labeling have proportioned the advancements in multi-modality. Various fields are increasingly using this form of learning to enhance applications, such as surveillance~\cite{chen2019distributed, alehdaghi2022visible, medeiros2024hallucidet}, industrial monitoring~\cite{kong2021deep, garillos2021multimodal, kini2023egocentric}, smart buildings~\cite{dubail2022privacy, dayarathna2023privacy}, self-driving cars~\cite{stilgoe2018machine, michaelis2019benchmarking, medeiros2024modality}, and robotics~\cite{eitel2015multimodal, pierson2017deep, ivorra2018multimodal}, due to their powerful ability to operate better in the presence of diverse environmental information~\cite{tang2023comparative}. For instance, the combination of visible (RGB) and infrared (IR) has been showing promising results regarding such applications due to the difference in light spectrum sensing by different sensors, which provide not only additional but also complementary information~\cite{wang2022improving}.

An unimodal learning (\fref{fig:cpirgb_different_learning}a), utilizes data from a single modality, for instance, an object detector trained and used in production with RGB images. In multimodal learning (\fref{fig:cpirgb_different_learning}b), the objective is to create a model able to incorporate information from multiple modalities, such as RGB and IR, from different sensors and requires paired modalities for both training and inference. Although this multimodal learning covers a wide range of applications, as aforementioned, we have identified an underserved scenario where one might want an RGB/IR modality agnostic model that is trained on both modalities but is subjected to only either one or another during inference (\fref{fig:cpirgb_different_learning}c). One example of that is a surveillance system where a server model is running all the time, and this model can provide detections for different RGB or IR sensors to address the need to make accurate detection in every lighting condition during different pre-defined conditions.

Despite the strong interest and business value in multimodal systems, much of the publicly available data and powerful pre-trained models are built around one modality: RGB. Furthermore, the lack of IR data gives additional motives to build a detector upon an already pre-trained unimodal RGB detector. However, the current methods proposed in research to incorporate dual-modality information into a model require dedicated components associated with each modality, which makes them incompatible with such RGB detectors. These methods are mainly fusion composed of techniques that adopt modalities by either distributing the modalities across a four-channel input (three RGB followed by one for IR), in the case of early fusion~\cite{Wagner2016}, merges both modalities later in the model architectures~\cite{9423251, 9191080, 10209020} in mid-stage fusion or ensembling different unimodal modality detectors~\cite{chen2022multimodal} in late-stage fusion. This constrains the model to utilize both modalities during inference, which significantly increases inference overhead compared to an unimodal architecture.

Typically, one probable phenomenon that can occur during a multimodal training is modality imbalance. This happens when the strongest modality is leveraged more than the others, leading to better overall performance while discarding contributions from the others.~\cite{das2023revisiting}. In this work, we provide a way to train a single shared vision encoder to be agnostic to its input RGB/IR modality yet still extract its knowledge during training to attain results almost as good on both modalities as if it was trained solely on each during testing/production. The naive solution for this type of task is to train a model with a dataset that blends both modalities.

Recently advances on patch-based transformers, such as ViT~\cite{DBLP:journals/corr/abs-2010-11929}, and Multi-Modal Masked Autoencoders~\cite{bachmann2022multimae} have steered us towards exploring patch-based architectures to build a powerful and yet simple training technique to create a RGB/IR modality agnostic vision encoder for object detection. Such approaches have been promising for multi-modal learning, which allows an efficient combination of different information~\cite{geng2022multimodal, bachmann2022multimae}. Our work investigates how to use RGB and IR modalities efficiently by using a patch-based transformer encoder. Thus, Mi(xed) Pa(tch) does not introduce any inference overhead during the testing phase while exploring an effective way to use the two modalities during the training. To accomplish such a task, we introduce a stochastic complementary patch mixing method, allowing the detector to explore each modality without having to rely on both of them simultaneously. This is possible by effectively sampling the optimal ratio of patches for each modality, which is then mixed using our technique. Subsequently, we enhance the training by suppressing the modality imbalances by proposing a modality-agnostic training technique, making the modalities indistinguishable from each other, a module inspired by Gradient Reversal Layer (GRL)~\cite{ganin2016domain} but with a novel design for patch based architectures. This approach is designed to allow low-cost inference in production while removing all requirements to know beforehand which modality the detector is going to be used with. Hence, in applications that run a detector all day, we can know beforehand that any of the modalities, RGB or IR, whenever they are being used, are going to perform optimally for the same shared vision encoder.

Our work provides empirical results alongside a theoretical explanation based on information theory describing the benefits of using MiPa with transformer-based backbones. Additionally, we study the ability of our MiPa to also be used as a regularization method for the more robust modality to boost the overall performance of the detector and we show that we can achieve competitive results on two traditional RGB/IR benchmarks: LLVIP and FLIR.

\noindent \textbf{Our main contributions can be summarized as follows:} \\
\noindent \textbf{(1)} We introduce MiPa, a novel mix patches RGB/IR modality agnostic training method for transformer-based object detectors, which learns how effectively sample the RGB and IR patches for best compressing the information of both modalities in a single encoder, without additional inference overhead. \\
\noindent \textbf{(2)} We propose a novel patch-wise modality agnostic module, which is inspired by the gradient reversal layer (GRL) for modality adaptation and is responsible to make the RGB/IR modalities invariant by the detector. \\
\noindent \textbf{(3)} We empirically demonstrate that the proposed method can also be used to improve the overall performance of detection when utilized as regularization for the strongest modality and achieve competitive results when compared with multimodal fusion methods, with less information during inference. Furthermore, MiPa can simply be applied to different transformer-based detectors, such as DINO~\cite{zhang2022dino} and Deformable DETR~\cite{zhu2020deformable}.

\section{Related Work}
\label{sec:related_work}

\noindent \textbf{Patch-Based Vision Encoding.} With the integration of Transformers in the vision field, researchers have started to deconstruct images into patches to allow the modeling of long-range relationships between patches~\cite{DBLP:journals/corr/abs-2010-11929}. This powerful approach yielded great results and quickly became the norm amongst the top-performing models, ranking well on popular benchmarks such as ImageNet-1k~\cite{DBLP:journals/corr/RussakovskyDSKSMHKKBBF14}. Multiple variants of the vision transformer have been proposed in recent years, for instance, ViT~\cite{DBLP:journals/corr/abs-2010-11929}, DEIT~\cite{DBLP:journals/corr/abs-2012-12877}, SWIN~\cite{liu2021swin}, and VOLO~\cite{DBLP:journals/corr/abs-2106-13112}. Alongside the new way of utilizing input images came a novel pretraining method for vision encoding: Masked Autoencoders~\cite{he2022masked} (MAE). Indeed, this technique, which is simple to understand and easy to implement, consists of using a classifier as an encoder in an autoencoder architecture to generate images by only using a small fraction of the patches as input. This unsupervised method has proven to be very useful in terms of improving results for downstream tasks. Furthermore, a similar idea has also been influential in the world of multi-modality models by building a multimodal MAE with one encoder and multiple decoders to reconstruct all the different modalities~\cite{bachmann2022multimae}. Recently, advances towards using SWIN Transformer as a backbone of DINO~\cite{zhang2022dino}, an object detector descendant of the DETR~\cite{carion2020end}, were responsible for reaching competitive results in detection benchmarks, such as in COCO dataset~\cite{lin2015microsoft}.

\noindent \textbf{Multimodal Visible-Infrared Object Detectors.} Regarding object detection, the primary methods of exploiting pairs of modalities, even when unaligned, are multimodal techniques; mainly fusion~\cite{Bayoudh2021ASO}. Fusion is a technique where the advantage of multiple modalities is taken in order to better optimize one training objective by combining them together to develop a multimodal representation~\cite{s23052381}. Fusion can be achieved at different stages, i.e., \textit{early-stage fusion}, which concatenates the modalities across the channels, \textit{mid-stage fusion}, where modalities are processed through dedicated decoders then merged e.g., Channel Switching and Spatial Attention (CSSA)~\cite{10209020}, Halfway Fusion~\cite{9191080}, RSDet~\cite{zhao2024removal}, CrossFormer~\cite{lee2024crossformer} or Guided Attentive Feature Fusion (GAFF)~\cite{9423251}, and finally \textit{late-stage fusion}, where typically modalities are processed independently through different models and combined at the end using ensembling~\cite{chen2022multimodal} e.g. ProbEn~\cite{chen2022multimodal}. The limitations of multimodal learning are that they require a custom architecture to handle each modality and are constrained to use both modalities during inference. A cross-modal with shared encoder vision models, however, are not affected by these limitations as the different modalities are only used during training and share the same encoder. This type of architecture unlocks the ability for detectors to have a higher degree of freedom for inference without compromising real-time applications. \\

\noindent \textbf{Modality Imbalance.} A potential obstacle to an RGB/IR modality-agnostic network is the phenomenon of modality imbalance. Given a dataset with multi-modal inputs, modality imbalance occurs when a model becomes more biased towards the contribution of one modality~\cite{das2023revisiting} than the others. To counter that, some methods have been proposed for classification, for instance, gradient modulation~\cite{9878839}, Gradient-Blending~\cite{9156420}, and Knowledge Distillation from the well-trained uni-modal model~\cite{du2021improving}. In gradient modulation, Peng et al. proposed a mechanism to control the adaptive optimization of each modality by monitoring their contributions to the learning objective. In gradient blending, Wang et al. identified that multi-modal learning can overfit due to the increased capacity of the networks and proposed a mechanism to blend the gradients effectively~\cite{9156420}. Du et al.~\cite{du2021improving} show that training multi-modal models on joint training can suffer from learning inferior representations for each modality because of the imbalance of the modalities and the implicit bias of the common objectives in the fusion strategy. An effective approach to help on the modality imbalance in a shared encoder consists of using a Gradient Reversal Layer (GRL)~\cite{ganin2015unsupervised}, which was introduced for domain adaptation to reduce a network’s reliance on a specific domain. GRL was exhaustively applied in object detection to create a shared domain; for instance, in~\cite{chen2018domain}, the GRL is used to adapt Faster R-CNN to distribution shifts in illumination or object appearance. The core idea of GRL involves training a classifier to identify the class of a data example during training. During backpropagation, the gradients are reversed to train the network to deceive the classifier.

In this work, we adapt this technique to address modality imbalance learning. Unlike typical cases where data belongs to a single domain/modality, a single training example of MiPa consists of a mosaic of the two modalities: RGB and IR. Therefore, our classifier is trained to predict a modality map instead. In our work, we tackle the imbalance with an adjustable balancing sampling, which learns how effectively sample the RGB and IR patches during training, and a patch-based GRL module responsible for encoding in the same vision encoder the information of both modalities while improving detection performance.

\begin{figure*}[!htp]
\centering
    \resizebox{\textwidth}{!}{%

    \begin{tabular}{c}
    
    \includegraphics[width=1.0\textwidth]{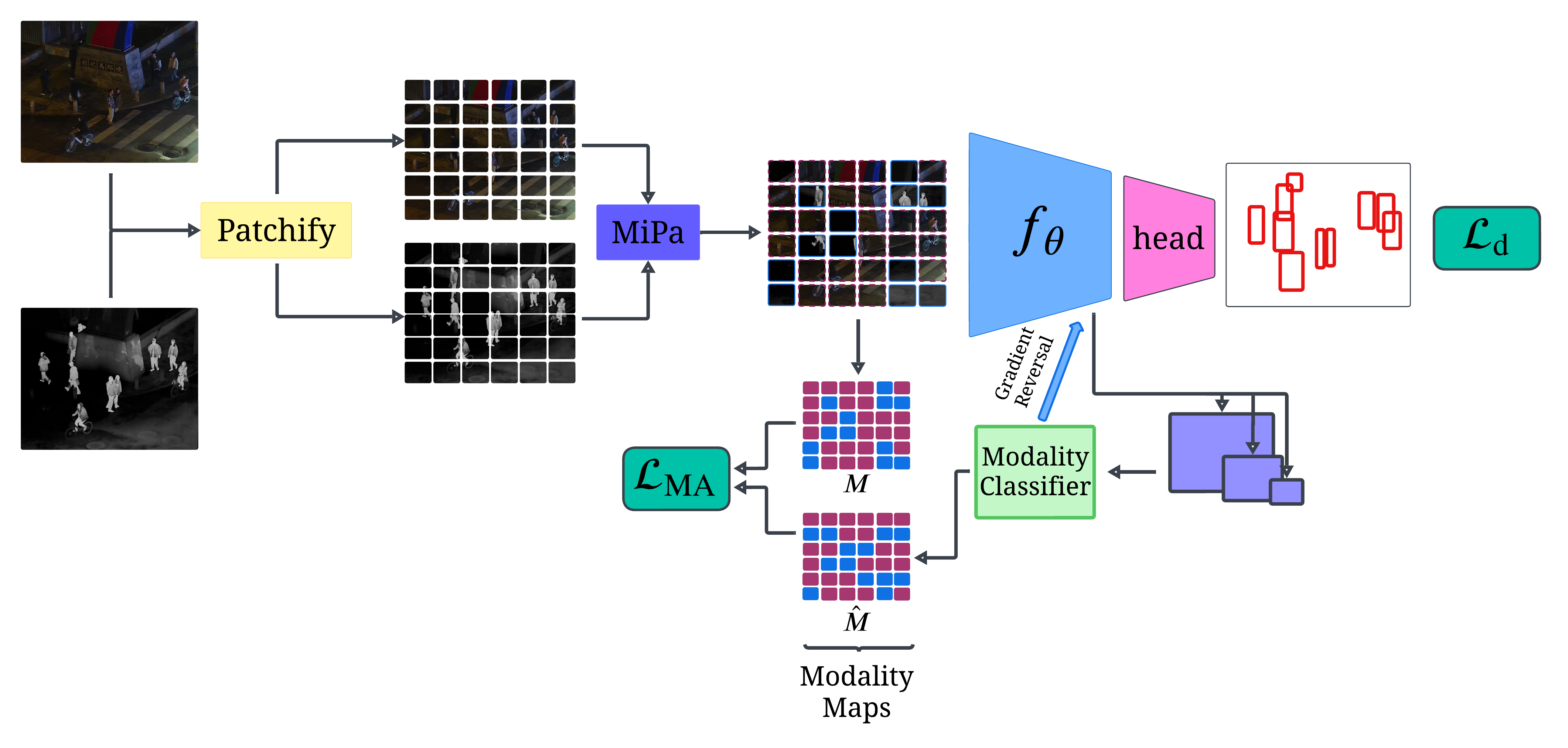} \\
    
    \end{tabular}
    }
\caption{Mixed Patches (MiPa) with Modality Agnostic (MA) module. In yellow is the patchify function. In purple is the MiPa module, followed by the feature extractor (encoder). In green is the modality classifier, and in pink is the detection head.}
\label{fig:cpirgb_general_mipa}
\end{figure*}

\section{Proposed Method} 
\label{sec:method}

While the naive way to create a multimodal vision encoder for an object detector is to blend the two modalities during training, we empirically show, in Section~\ref{sec:experiments}, that this approach does not lead to a balanced performance on both modalities. In this section, we present our proposed solution.

\subsection{Preliminary definitions.}

Let us consider a set of training samples $\mathcal{D} = \{(x_{i}, B_{i})\}$ where $x_{i} \in \mathbb{R}^{W \times H \times C}$ is the image $i$ with spatial resolution $W \times H$ and $C$ channels. Here, a set of bounding boxes is represented by $B_{i} = \{{b_{0}, b_{1},...,b_{N}\}}$ with $b = (c_{x}, c_{y}, w, h)$ being $c_{x}$ and $c_{y}$ coordinates of the center of the bounding box with size $w \times h$. During the training process of a neural network-based detector, we aim to learn a parameterized function $f_{\theta}: \mathbb{R}^{W \times H \times C} \rightarrow \mathcal{B}$, being $\mathcal{B}$ the family of sets $B_{i}$ and $\theta$ the parameters vector. For such, the optimization is guided by a loss function, which is a combination of
a regression $\mathcal{L}_{r}$ and a classification $\mathcal{L}_{c}$ term, i.e., $l_{2}$ loss and binary cross-entropy, respectively. The following Equation~\eqref{eq:detection_loss} defines a general loss function ($\mathcal{L}_{d}$) for object detection:

\begin{equation}
   \mathcal{L}_{d}(\theta)=\frac{1}{|\mathcal{D}|} \sum_{(x, B) \in \mathcal{D}} \mathcal{L}_{c}(f_{\theta}(x), B)+\lambda \mathcal{L}_{\text {r}}(f_{\theta}(x), B).
    \label{eq:detection_loss}
\end{equation}

\subsection{Mixed Patches (MiPa).}

The MiPa training method is a training technique that leverages the patch input channel from transformer-based feature extractors to build a powerful common representation between RGB/IR modalities for the unique vision encoder, which can be used in different transformer-based detectors. In short, it consists of a single encoder that receives sampling complementary patches from each modality and rearranges the input into a sort of mosaic image as shown in \fref{fig:cpirgb_general_mipa}. Such mechanism forces the model to see both modalities for each inference without being forced to have parameters specialized on a specific one. Depending on how the nature of the patches are sampled, the technique can act as a way to gather the union of information between both modalities or as a regularization for the strongest modality (the easiest modality that tends to drive the learning process). Throughout this paper, we will reference the sampling ratio of the patches as \(\rho\). There are several ways to pick the sampling ratio \(\rho\); the naive way of selecting \(\rho\) is to use a fixed ratio during the training of $50\%$. Then, we can randomly generate a \(\rho\) value for each inference. If we have an intuition of which modality needs to be sampled more, we can manually move \(\rho\) during the training with a certain curriculum. Finally, we can let the model learn the optimal ratio by itself. In this work, we have explored all these variations to see which one is the most suitable for MiPa.

\begin{table}[!htp]
    
    \caption{Definition of the random variables and information measures used to explain MiPa.}    \centering
    \resizebox{1.0\columnwidth}{!}{%
    
    \begin{tabular}{qq}
    \toprule
       \rowcolor{white}

        \multicolumn{2}{c}{\rule{0pt}{4ex}\textbf{General}} \\ \rowcolor{white} \\

        \midrule

        Input scene in patches & $\mathcal{X}_n$ \\
        \midrule
        \rowcolor{white}
        Number of patches & $n$ $\in$ $\mathbb{N}$ \\
        \midrule
        Patch id & $i$ $\in$ $\mathbb{N}$ \\
        \bottomrule

        \rowcolor{white}
        \multicolumn{2}{c}{\rule{0pt}{4ex}\textbf{Random variables (RVs)}} \\ \rowcolor{white} \\

        \toprule
        Patch ratio & $\rho \sim U(0,1)$ \\
        \midrule 
        \rowcolor{white}
        Patch channel $f$ &  $m \sim \binom{n\cdot\rho} {p}$, p = $\frac{1}{2}$ \\
        \midrule 
        Patch channel $g$ &  $l \sim n$ - $m$ \\
        \bottomrule

        \multicolumn{2}{c}{\rule{0pt}{4ex}\textbf{Functions}} \\ \rowcolor{white} \\

        \toprule
        MiPa & $\mathcal{M}$ \\
        \midrule
        \rowcolor{white}
        Self-Attention & $SA$ \\
        \midrule
        Modality channels &  $f$, $g$ \\
        \bottomrule

        \multicolumn{2}{c}{\rule{0pt}{4ex}\textbf{Information measures}} \\ \rowcolor{white} \\
        
        \toprule
        Entropy of $V$ & $ \mathcal{H}({V})\coloneqq \mathbb{E}_{p_{V}}\left[-\log p_V (V)\right]$ \\
        \midrule
        \rowcolor{white}
        Information of \textit{$X$}  & $Q, P$ where $Q$ =\textit{$\mathcal{H}(q)$}, $P$ = \textit{$\mathcal{H}(p)$}  \\
        \midrule
        Noise modality channels & $\eta$  \\
        \midrule
        \rowcolor{white}
        Mutual information between $P$ and $Q$ & \textit{$I(Q, P)$} = \textit{$Q$} + \textit{$P$} - \textit{$Q$}$\cap$\textit{$P$}  \\
        \midrule
        Approximation of mutual information between $P$ and $Q$ & \textit{$I_a$} $\approx$ \textit{$I$}  \\

    \bottomrule
   
    \end{tabular}
    }

\label{table:notations}
\end{table}

\textbf{Theoretical explanation behind the MiPa approach.}  Here, we detail our theoretical understanding of the MiPa method. We refer to~\tref{table:notations} for all definitions. The variable $\mathcal{X}$ can be thought of as a scene where you would see individuals walking in the street, for instance, and the functions $f$ and $g$ are camera lenses capturing the information of the scene via IR and RGB, respectively. The goal of MiPa ($\mathcal{M}$) is to enhance learning efficiency by merging information from both modalities, eliminating redundancy, and filtering out noise, all in a \textbf{single inference}.\\
\noindent \textit{Thus, say we have:}

\begin{equation}
f(\mathcal{X}) = P + \eta_f; g(\mathcal{X}) = Q + \eta_g,
\label{eq:f_and_g}
\end{equation}

\noindent where Equation~\eqref{eq:f_and_g} represents the visualization of the scene, which is composed of the information captured ($P$ or $Q$) by the sensor and some noise ($\eta$). Then the application of MiPa ($\mathcal{M}$) can be summarized as the following Equation~\eqref{eq:mipa}:

\begin{equation}
\begin{aligned}
\mathcal{M}(f(\mathcal{X}), g(\mathcal{X})) &
= \begin{cases}
    f(\mathcal{X}_i),& i\in m\\
    g(\mathcal{X}_i),& i\in l,
\end{cases} \\
\end{aligned}
\label{eq:mipa}
\end{equation}

\noindent where $f(\mathcal{X}_i)$ represents the mapping of the patch $\mathcal{X}$ with id $i$ using $f$ (IR lens) and $g(\mathcal{X}_i)$ using RGB lens. Then, the combination of the individual patches of each modality is given by Equation~\eqref{eq:mipa_combination}:

\begin{equation}
\mathcal{M} = (P_0 + P_1 + Q_2 + ... + Q_{n-1} + P_n) + (m\cdot\eta_f + l\cdot\eta_g).
\label{eq:mipa_combination}
\end{equation}

\noindent As RGB and IR patches do not encode the same information in the same patch visualization $\mathcal{X}_i$, the additional information of one modality improves, for instance, IR on the night, the other one. Also, this variation in the sense of information for both modalities is also responsible for regularizing the training when the patches are mixed. The following Equation~\eqref{eq:total_information} represents the approximation of the real mutual information $I$ by $\mathcal{M}$ using Equation~\eqref{eq:mipa_combination} and approximating the noise from the scene to be similar for both sensors:

\begin{equation}
\mathcal{M} = I_a + \eta.
\label{eq:total_information}
\end{equation}

\noindent This approximation means that the encoded information on MiPa represents the total scene composed by both sensors, which are compressed on the vision encoder while removing the redundancy information and noise by the training process.

\subsection{Patch-Wise Modality Agnostic Training.}
As previously mentioned, modality imbalances can potentially cause the model to rely mostly on one modality. Since the objective of this work is to preserve the original architecture of the model for inference, we opted for an approach where the backbone would have the responsibility of mediating the modalities. To do so, we designed an adaption of the GRL technique~\cite{ganin2015unsupervised} called \textbf{\textit{patch-wise modality agnostic}} module. The key idea is to prevent the detector from relying too much on the strongest modality, the easiest modality driven by the learning process, by making the features from each modality indistinguishable, therefore sharing the same encoding. Considering that the input has a different modality for each patch, a modality that we pick during the patch mixing process, we build what we call a \textit{modality map}, denoted as $M$, that specifies which modality each patch belongs to for each inference during training. Then, we use a modality classifier to predict the modality map of the features coming from the backbone. Finally, we compute the loss between the target and outputted modality maps and back-propagate the opposite gradients to the backbone encoder. To reduce the noise coming from the classifier at the beginning of the training, we slowly increase the weight ($\lambda$) of the gradients propagated to the backbone as the training goes on. We use the Binary Cross-Entropy (BCE) to compute the loss between the predicted and target modality maps, as described by the following  Equation~\eqref{eq:modality_invariance_bce}:

\begin{equation}
    \mathcal{L}_{\text{MA}} = \frac{1}{n}\sum_{i=1}^{n}-{M\log(\hat{M}) - (1 - M)\log(1 - \hat{M})},
    \label{eq:modality_invariance_bce}
\end{equation}

\noindent where $M$ is the modality map generated from $\rho$. The aforementioned approach for the full training pipeline can be seen in~\fref{fig:cpirgb_general_mipa}. We use the following  Equation~\eqref{eq:modality_invariance_loss} to increment the factor $\lambda$.

\begin{equation}
    \lambda = \frac{2}{1 + exp(-\gamma · s)} - 1,
    \label{eq:modality_invariance_loss}
\end{equation}

\noindent where \textbf{$\gamma$} is the speed to which \textbf{$\lambda$} increases. The modality classifier can be used at any stage of the backbone; we have found empirically that using it on the features from the stage $1$ works well. Finally, MiPa loss ($\mathcal{L}_{\text{MIPA}}$) can be defined as the following Equation~\eqref{eq:mipa_loss}:

\begin{equation}
    \mathcal{L}_{\text{MIPA}} = \mathcal{L}_{d} + \lambda\mathcal{L}_{\text{MA}}.
    \label{eq:mipa_loss}
\end{equation}

\section{Results and Discussion}
\label{sec:experiments}

\subsection{Experimental Setup and Methodology}

\textbf{Datasets.} During our experiments, we explored two different RGB/IR benchmarking datasets: LLVIP and FLIR. \\
\textbf{LLVIP:} The LLVIP dataset is a surveillance dataset composed of $12,025$ IR and $12,025$ RGB paired images for training and $3,463$ IR and $3,463$ RGB paired images for testing. The original resolution is $1280$ by $1024$ pixels but was resized to $640$ by $512$ to accelerate the training. The sole annotated class of this dataset is pedestrians. \textbf{FLIR ALIGNED:} For the FLIR dataset, we used the sanitized and aligned paired sets provided by Zhang et al.~\cite{zhang2020multispectral}, which has $4,129$ IRs and $4,129$ RGBs for training, and $1,013$ IRs and $1,013$ RGBs for testing. The FLIR images are taken from the perspective of a camera in the front of a car, and the resolution is $640$ by $512$. It contains annotations of bicycles, dogs, cars, and people. It has been found that for the case of FLIR, the "dog" objects are inadequate for training~\cite{10209020}, but since our objective is to evaluate if our method can make a detector modality agnostic and not beat any prior benchmark, we have decided to keep it during our evaluations. \newline

\textbf{Implementation details.} All our detectors were trained on an A100 NVIDIA GPU and were implemented using PyTorch. We use AdamW~\cite{loshchilov2017decoupled} as an optimizer with a learning rate of $1e^{-4}$, a batch size of $6$, and for a total of $12$ epochs for the case of the DINO~\cite{zhang2022dino} detector. For SWIN, we start with the pre-trained weights from ImageNet \cite{DBLP:journals/corr/RussakovskyDSKSMHKKBBF14}. We evaluated the models in terms of performance AP$_{50}$, and we additionally reported the others AP$_{75}$ and AP in the supplementary material. Furthermore, the evaluation is done in terms of RGB performance, IR performance, and our target metric, the average of both, because our setup requires a model that is equally good on both modalities during test time.

\begin{figure*}
\centering
\begin{subfigure}[t]{0.24\textwidth}
    \caption{RGB - Night}

    \makebox[0pt][r]{\makebox[15pt]{\raisebox{30pt}{\rotatebox[origin=c]{90}{GT}}}}%
    \includegraphics[width=\textwidth]{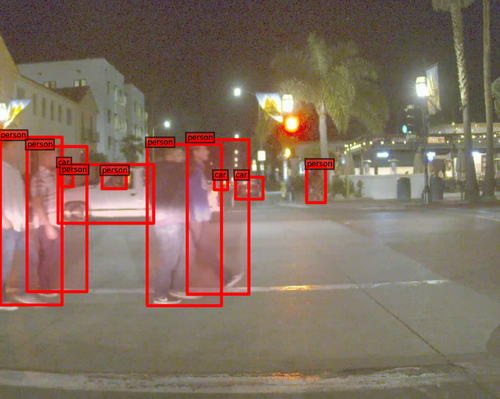}

    \makebox[0pt][r]{\makebox[15pt]{\raisebox{30pt}{\rotatebox[origin=c]{90}{RGB}}}}%
    \includegraphics[width=\textwidth]{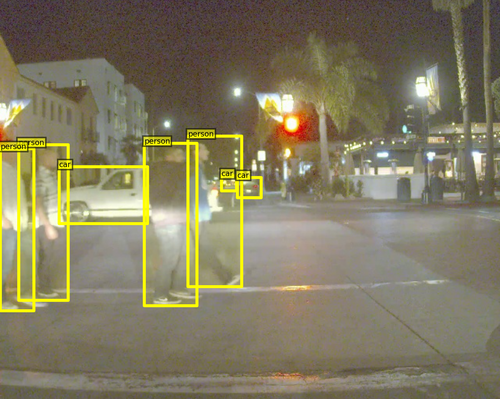}

    \makebox[0pt][r]{\makebox[15pt]{\raisebox{30pt}{\rotatebox[origin=c]{90}{IR}}}}%
    \includegraphics[width=\textwidth]{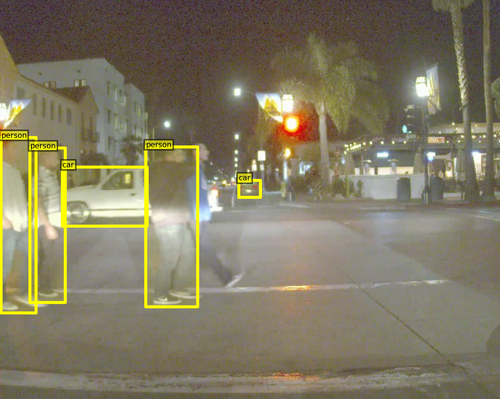}

    \makebox[0pt][r]{\makebox[15pt]{\raisebox{30pt}{\rotatebox[origin=c]{90}{Both}}}}%
    \includegraphics[width=\textwidth]{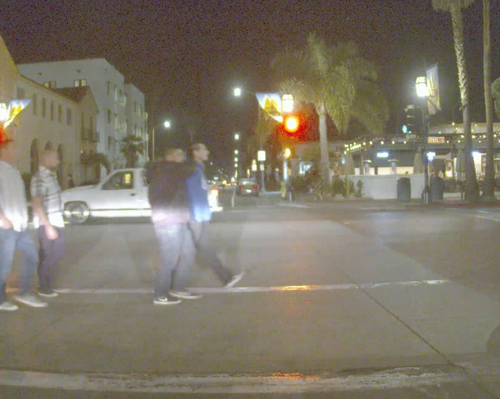}

    \makebox[0pt][r]{\makebox[15pt]{\raisebox{30pt}{\rotatebox[origin=c]{90}{MiPa}}}}%
    \includegraphics[width=\textwidth]{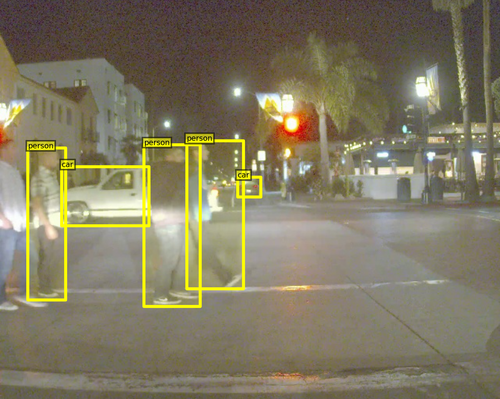}

\end{subfigure}
\begin{subfigure}[t]{0.24\textwidth}
    \caption{IR - Night}

    \includegraphics[width=\textwidth]{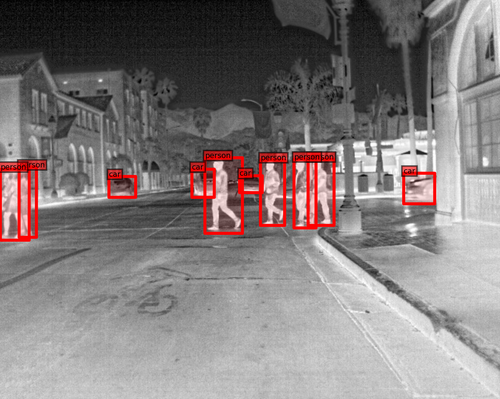}

    \includegraphics[width=\textwidth]{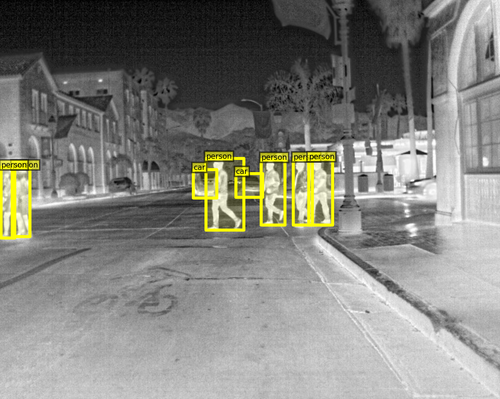}
    
    \includegraphics[width=\textwidth]{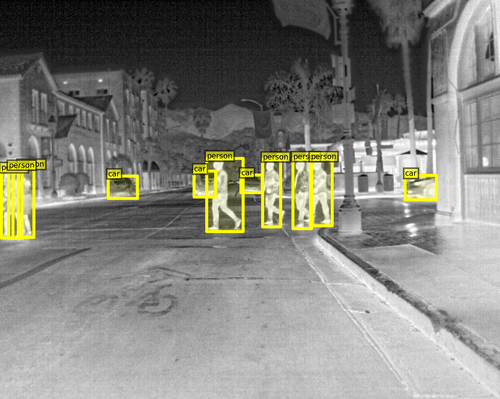}

    \includegraphics[width=\textwidth]{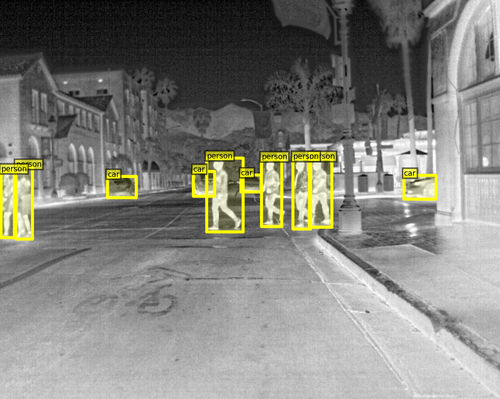}

    \includegraphics[width=\textwidth]{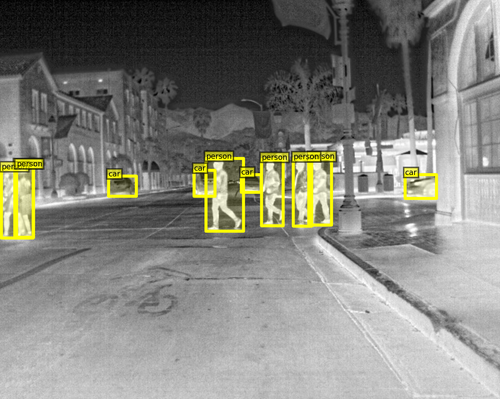}

\end{subfigure}
\begin{subfigure}[t]{0.24\textwidth}
    \caption{RGB - Day}
    
    \includegraphics[width=\textwidth]{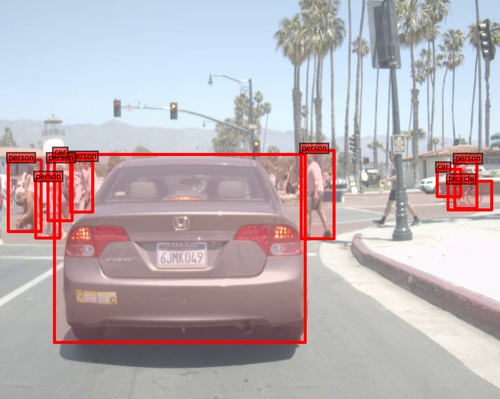}
    
    \includegraphics[width=\textwidth]{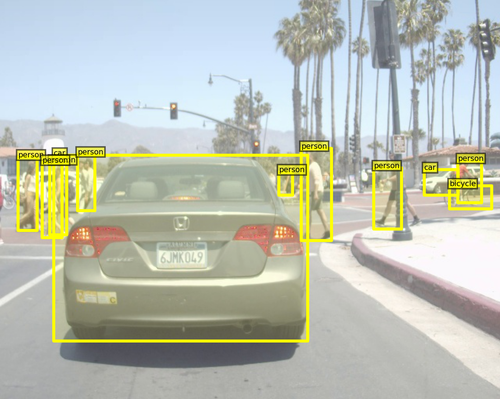}

    \includegraphics[width=\textwidth]{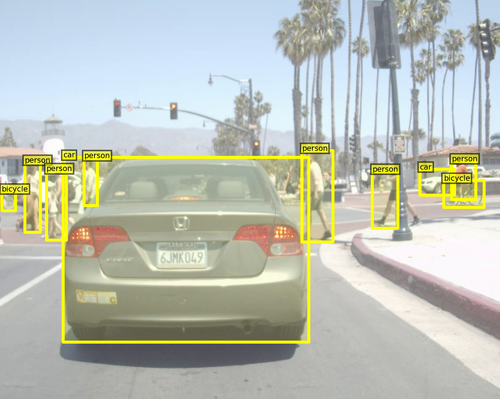}
    
    \includegraphics[width=\textwidth]{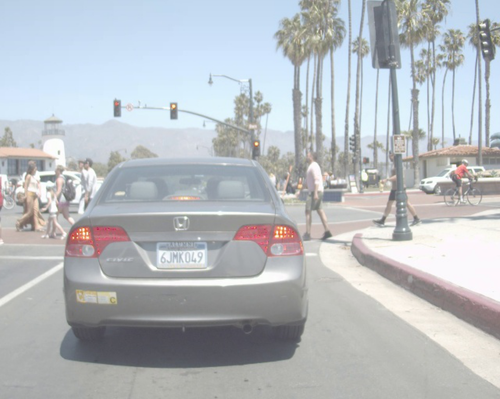}

    \includegraphics[width=\textwidth]{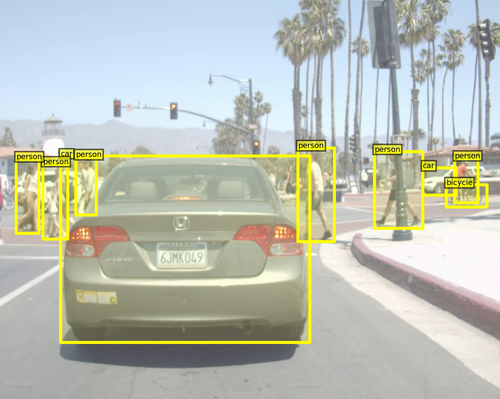}

\end{subfigure}
\begin{subfigure}[t]{0.24\textwidth}
    \caption{IR - Day}

    \includegraphics[width=\textwidth]{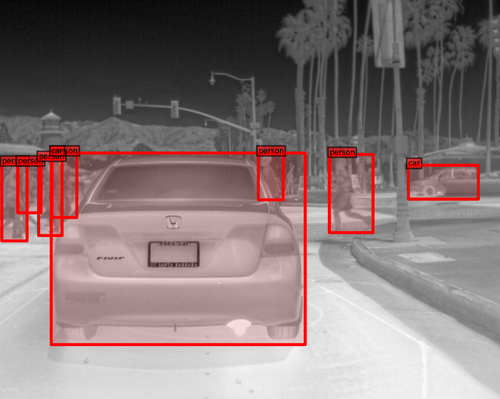}
    
    \includegraphics[width=\textwidth]{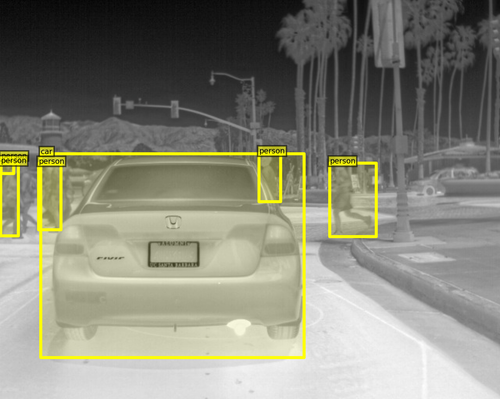}

    \includegraphics[width=\textwidth]{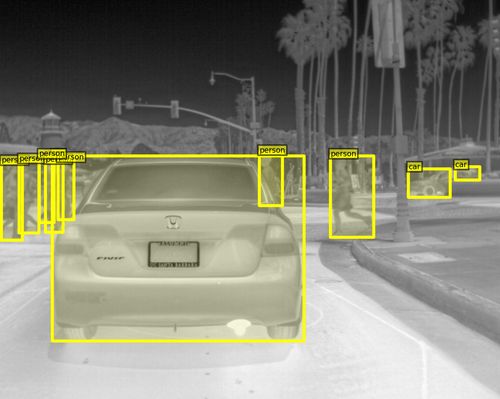}
    
    \includegraphics[width=\textwidth]{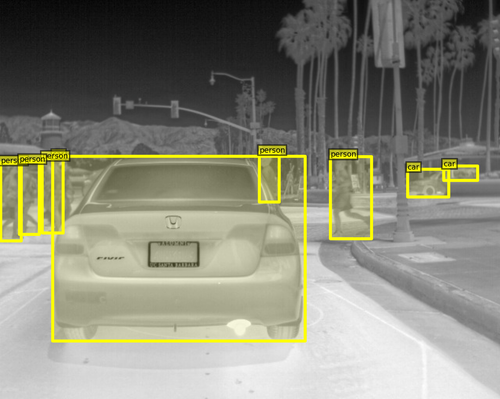}

    \includegraphics[width=\textwidth]{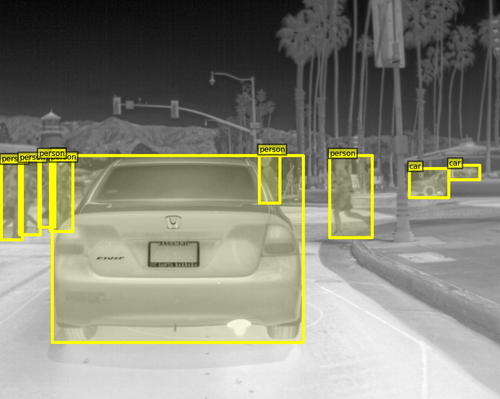}

\end{subfigure}

\caption{Detection over different methods for two different daytimes: Night and Day and two different modalities: RGD and IR. Detectors trained on \textit{RGB} work better in the daytime. Detectors trained on \textit{IR} work better at nighttime. Detectors trained on \textit{Both} modalities in a naive way cannot work only on the dominant modality. Our \textit{MiPa} manages to work well in all conditions.}
\label{fig:daynight}
\end{figure*}

\subsection{Baselines} 
In the course of this work, we considered different baselines to compare to our proposed method (MiPa). Firstly, we measure the performance of the detector trained on one modality, uni-modal setup, to gain a reference of the expected detection coming from each modality. Secondly, we evaluate the naive solution of simply using a dataset comprised of both modalities during training (multimodal setting), which we call~\textit{Both}. To account for the modality imbalances and further increase the fairness of our comparisons, we balanced the datasets with $25\%$, $50\%$, and $75\% $ of one modality and the rest of the other. All models were evaluated separately on RGB and IR. Additionally, the mean of the modalities, which represents how well the model is balanced for the two desired modalities, is calculated.

\begin{table}[!htp]
    
    \caption{Comparison of different ratio  \(\rho\) sampling methods on LLVIP. Using DINO with SWIN backbone.}
    \centering
    \resizebox{0.9\columnwidth}{!}{%
    \begin{tabular}{gggg}
    \toprule
       \rowcolor{white}
         {} & \multicolumn{3}{c}{\textbf{Dataset: LLVIP} (\textbf{AP$_{50}\uparrow$} )} \\
        
        \cmidrule(lr){2-4}
        \addlinespace[5pt]
                
        \cellcolor[HTML]{FFFFFF} \multirow{1}{*}[+0.9em]{\textbf{Model}} &  \cellcolor[HTML]{FFFFFF}\multirow{2}{*}[1em]{\textbf{RGB}} & \cellcolor[HTML]{FFFFFF}\multirow{2}{*}[1em]{\textbf{IR}} & \cellcolor[HTML]{FFFFFF}\multirow{2}{*}[1em]{\textbf{Average}}   \\

        \midrule
       \rowcolor{white}
        Fixed [\(\rho\)=0.25] & 78.9    & \textbf{98.2}    & 88.55 \\																			
        Fixed [\(\rho\)=0.50] & 73.0    & 97.6    & 85.30 \\												
        \rowcolor{white}
        Fixed [\(\rho\)=0.75] & 77.4    & 97.5    & 87.45 \\

        \midrule
        Curriculum (\(\rho\)=0.25 / $4$ epochs)  & 76.6    & 97.8    & 87.20 \\
        
        \rowcolor{white}
        Curriculum (\(\rho\)=0.25 / $8$ epochs) & 80.1    & 97.8    & 88.95 \\

        \midrule
        Variable    & \textbf{88.5}    & 97.5    & \textbf{93.00} \\

    \bottomrule
   
    \end{tabular}
    }

\label{tab:all_rho}
\end{table}

\subsection{Towards the optimal $(\rho)$}

Since the way of selecting the ideal \(\rho\) was not clear, we designed different experimental settings to study the influence of \(\rho\) on learning the best way to balance the amount of RGB/IR information during the training. Let us start with a few definitions. \newline 

\noindent \textbf{Fixed \(\rho\).} In the fixed setting, we selected a fixed amount of proportion between RGB/IR sampling, such as $0\%$, $25\%$, $50\%$, $75\%$ and $100\%$, in which $0\%$ correspond to none IR at each batch, and $100\%$ correspond to only IR on the training batch. \newline

\noindent \textbf{Curriculum \(\rho\).} For the curriculum strategy, we tested different times during the training to give different importance to one modality over the other. For instance, during the initial epochs over the training, the model focuses on the easier-to-learn modality (IR modality tends to drive the learning process when a balanced jointly dataset is given), providing between $0\%$ to $25\%$ of ratio for IR, and then over the rest of training, it samples from the uniform distribution such as variable \(\rho\). \newline

\noindent \textbf{Variable \(\rho\).} In the variable \(\rho\), the ratio of mixed patches per batch is drawn from a uniform distribution. For each batch, a different \(\rho\) is redrawn. \newline

\noindent We tested all the different configurations of \(\rho\) on LLVIP (see~\tref{tab:all_rho}). For this experiment, we have made two findings. First, using an $I_a$ following a uniform distribution gives us a better approximation of the range of information from $IR\cup RGB$ as the results from the variable give us a better balance between both modalities. Second, using less of the weaker modality (hard to learn) strengthens the learning of the strongest one (easier to learn modality), as it can be seen in our table that we were actually able to beat the state-of-the-art by sampling $25\%$ of RGB images and $75\%$ of IR.

\begin{table}[!htp]
    
    \caption{Comparison of detection performance over different baselines and MiPa for different models on SWIN backbone for DINO and Deformable DETR. The evaluation is done for RGB, IR, and the average of the modalities.}
    \centering
    \resizebox{1.0\columnwidth}{!}{%
    \begin{tabular}{lqggg}
    \toprule
       \rowcolor{white}
        {} & {} & \multicolumn{3}{c}{\textbf{Dataset: LLVIP} (\textbf{AP$_{50}\uparrow$} )} \\
        
        \cmidrule(lr){3-5}
        \addlinespace[5pt]
                
        \cellcolor[HTML]{FFFFFF} \multirow{1}{*}[+0.9em]{\textbf{Detector}} &
        \cellcolor[HTML]{FFFFFF} \multirow{1}{*}[+0.9em]{\textbf{Model}} &  \cellcolor[HTML]{FFFFFF}\multirow{2}{*}[1em]{\textbf{RGB}} & \cellcolor[HTML]{FFFFFF}\multirow{2}{*}[1em]{\textbf{IR}} & \cellcolor[HTML]{FFFFFF}\multirow{2}{*}[1em]{\textbf{Average}}   \\
        
        \midrule
       \rowcolor{white}
        {} & RGB & 90.87 ± 0.84 &  94.23 ± 0.57  &	92.55 \\																						
        {} &  IR & 66.87 ± 0.90 &	96.87 ± 0.12 & 81.87 \\													
        \rowcolor{white}
        \multirow{1}{*}[-1.2em]{\textbf{DINO}} & Both [$\rho=0.25$] & 79.73 ± 1.03 & 97.40 ± 0.22  &	88.57 \\
        {} & Both [$\rho=0.50$] & 82.40 ± 1.50 & 96.50 ± 0.29  &	89.45 \\
        
        \rowcolor{white}
        {} & Both [$\rho=0.75$]   & 81.23 ± 2.89 & 97.07 ± 0.25  &	89.15 \\
        
        \cmidrule(lr){2-5}
        {} & \textbf{MiPa (Ours)} &  88.70 ± 0.45 & 96.97 ± 0.26   &\textbf{92.83} \\

        {} & \textbf{MiPa + MA (Ours)} & 89.10 ± 0.28 &	96.83 ± 0.09   &	\textbf{92.90} \\

        \midrule

        \rowcolor{white}        
        {} & RGB  & 80.00 ± 1.50 & 90.03 ± 00.87 & 85.02 \\
        
        {} & IR & 56.10 ± 2.50 & 94.20 ± 00.08 & 75.15 \\
        
        \rowcolor{white}
        \multirow{1}{*}[-1.2em]{\textbf{Def.DETR}}  & Both [$\rho=0.25$]  & 51.20 ± 3.47 & 83.73 ± 16.57 & 67.47 \\															
        {} & Both [$\rho=0.50$]  & 53.57 ± 4.17 & 83.87 ± 16.17 & 68.72 \\	
        
        \rowcolor{white}
        {} & Both [$\rho=0.75$] & 53.53 ± 4.55  & 82.33 ± 18.48 & 67.93 \\

        \cmidrule(lr){2-5}
        {} & \textbf{MiPa (Ours)} &  78.60 ± 0.42 &	95.20 ± 0.16 &	\textbf{86.90}  \\																		
        {} & \textbf{MiPa + MA (Ours)} &  79.02 ± 0.21 &	95.36 ± 0.25  &	\textbf{87.19} \\

        \midrule

       \rowcolor{white}
        {} & {} & \multicolumn{3}{c}{\textbf{Dataset: FLIR} (\textbf{AP$_{50}\uparrow$} )} \\
        
        \cmidrule(lr){3-5}
        \addlinespace[5pt]
                
        \cellcolor[HTML]{FFFFFF} \multirow{1}{*}[+0.9em]{\textbf{Detector}} &
        \cellcolor[HTML]{FFFFFF} \multirow{1}{*}[+0.9em]{\textbf{Model}} &  \cellcolor[HTML]{FFFFFF}\multirow{2}{*}[1em]{\textbf{RGB}} & \cellcolor[HTML]{FFFFFF}\multirow{2}{*}[1em]{\textbf{IR}} & \cellcolor[HTML]{FFFFFF}\multirow{2}{*}[1em]{\textbf{Average}}   \\
        
        \midrule
        
       \rowcolor{white}
        {} & RGB & 66.07 ± 0.98 &	56.60 ± 0.80 &	61.33 \\

        {} & IR & 56.47 ± 0.79 &	70.40 ± 0.38 & 63.43 \\
        
        \rowcolor{white}
        \multirow{1}{*}[-1.2em]{\textbf{DINO}} & Both [$\rho=0.25$]  &  56.53 ± 0.76 & 67.57 ± 1.73 & 62.05 \\		
        
        {} & Both [$\rho=0.50$]  & 60.50 ± 0.66 & 68.93 ± 0.60 &	64.72 \\	
        
        \rowcolor{white}
        {} & Both [$\rho=0.75$]   & 58.53 ± 0.92 & 70.43 ± 0.65 & 64.48 \\	
        
        \cmidrule(lr){2-5}
        {} & \textbf{MiPa (Ours)} & 63.53 ± 1.94 & 69.50 ± 1.84 & \textbf{66.52} \\												
        {} & \textbf{MiPa + MA (Ours)} & 64.80 ± 2.30 & 70.43 ± 0.53  & \textbf{67.62} \\

        \midrule

        \rowcolor{white}        
        {} & RGB  & 49.33 ± 1.39 &  43.77 ± 00.56 &  46.55  \\
        
        {} & IR & 39.17 ± 1.48 & 59.20 ± 00.29 & 49.18 \\
        
        \rowcolor{white}
        \multirow{1}{*}[-1.2em]{\textbf{Def.DETR}}  & Both [$\rho=0.25$]  & 35.73 ± 4.95 & 43.00 ± 13.54 & 39.37 \\												
        {} & Both [$\rho=0.50$]  &  33.93 ± 5.15 & 43.33 ± 14.14 & 38.63 \\
        
        \rowcolor{white}
        {} & Both [$\rho=0.75$] & 32.90 ± 3.54 & 44.13 ± 14.85 & 38.52 \\

        \cmidrule(lr){2-5}
        {} & \textbf{MiPa (Ours)} & 48.00 ± 0.57 & 54.97 ± 00.90 & \textbf{51.48} \\															
        {} & \textbf{MiPa + MA (Ours)} & 48.27 ± 1.76 & 55.80 ± 00.22 & \textbf{52.03} \\

    \bottomrule

    \end{tabular}
    }

\label{tab:tab0_swin}
\end{table}

\begin{table}[!htp]
    \caption{MiPa ablation on $\gamma$ and comparison with different baselines for DINO SWIN. The evaluation is done for RGB, IR, and the average of the modalities in terms of AP$_{50}$ performance.}
    \centering
    \resizebox{1.0\columnwidth}{!}{%
    \begin{tabular}{qgggggggggggggg}

        \toprule

       \rowcolor{white}
        {} & \multicolumn{3}{c}{\textbf{Dataset: LLVIP} (\textbf{AP$_{50}\uparrow$} )} \\
        
        \cmidrule(lr){2-4}
        \addlinespace[5pt]
                
        \cellcolor[HTML]{FFFFFF} \multirow{1}{*}[+0.9em]{\textbf{Modality}} &  \cellcolor[HTML]{FFFFFF}\multirow{2}{*}[1em]{\textbf{RGB}} & \cellcolor[HTML]{FFFFFF}\multirow{2}{*}[1em]{\textbf{IR}} & \cellcolor[HTML]{FFFFFF}\multirow{2}{*}[1em]{\textbf{Average}}   \\
        
        \midrule
        
       \rowcolor{white}
        RGB & 90.87 ± 0.84 & 94.23 ± 0.57 & 92.55 \\

        IR & 66.87 ± 0.90 & 96.87 ± 0.12 & 81.87 \\
        
        \rowcolor{white}
        Both [$\rho=0.25$]  & 79.73 ± 1.03 & 97.40 ± 0.22 & 88.57 \\
        
        Both [$\rho=0.50$] & 82.40 ± 1.50 & 96.50 ± 0.29 & 89.45 \\
        
        \rowcolor{white}
        Both [$\rho=0.75$]  & 81.23 ± 2.89 & 97.07 ± 0.25 & 89.15 \\
        
        \midrule
        MiPa & 88.70 ± 0.45 & 96.97 ± 0.26 & 92.83 \\
		
        \rowcolor{white}
        MiPa [$\gamma=0.05$] & 89.20 ± 0.43 & 96.57 ± 0.39 & 92.88 \\				
        
        MiPa [$\gamma=0.10$] & 89.43 ± 0.25 & 96.57 ± 0.31 & \textbf{93.00} \\				
        
        \rowcolor{white}
        MiPa [$\gamma=0.15$] & 89.10 ± 0.28 & 96.83 ± 0.09 & 92.97 \\

        \midrule
       \rowcolor{white}
        {} & \multicolumn{3}{c}{\textbf{Dataset: FLIR} (\textbf{AP$_{50}\uparrow$} )} \\
        
        \cmidrule(lr){2-4}
        \addlinespace[5pt]
                
        \cellcolor[HTML]{FFFFFF} \multirow{1}{*}[+0.9em]{\textbf{Modality}} &  \cellcolor[HTML]{FFFFFF}\multirow{2}{*}[1em]{\textbf{RGB}} & \cellcolor[HTML]{FFFFFF}\multirow{2}{*}[1em]{\textbf{IR}} & \cellcolor[HTML]{FFFFFF}\multirow{2}{*}[1em]{\textbf{Average}}   \\
        
        \midrule
        
       \rowcolor{white}
        RGB & 66.07 ± 0.98 &	56.60 ± 0.80  &	61.33 \\

        IR & 56.47 ± 0.79 &	70.40 ± 0.38 &	63.43 \\
        
        \rowcolor{white}
        Both [$\rho=0.25$]  &  56.53 ± 0.76 &	67.57 ± 1.73 & 62.05 \\
        
        Both [$\rho=0.50$]  & 60.50 ± 0.66  &	68.93 ± 0.60 &	64.72 \\
        
        \rowcolor{white}
        Both [$\rho=0.75$]   & 58.53 ± 0.92  &	70.43 ± 0.65 &	64.48 \\
        
        \midrule
        MiPa & 63.53 ± 1.94 & 69.50 ± 1.84  &	66.52 \\
	
        \rowcolor{white}
        MiPa [$\gamma=0.05$] & 64.80 ± 2.30  &	70.43 ± 0.53 &	\textbf{67.62} \\													
        
        MiPa [$\gamma=0.10$] & 64.03 ± 2.11   &	69.63 ± 1.45  &	66.83 \\																						
        \rowcolor{white}
        MiPa [$\gamma=0.15$] & 64.27 ± 0.47  &	69.93 ± 1.02 &	67.10 \\															
    \bottomrule
        
    \end{tabular}
    }
\label{tab:ablation_gamma}
\end{table}

\subsection{Patch-wise Modality Agnostic Training.} 

The subsequent ablation shows the efficacy of the patch-wise modality agnostic method towards obtaining a single model capable of dealing with both modalities while keeping the performance stable. In~\tref{tab:ablation_gamma}, we studied the sensibility of the model performances influenced by different $\gamma$ hyperparameters, seen in Equation~\eqref{eq:modality_invariance_loss}, which tunes the speed to which the $\lambda$ factor increases at each step the weight of gradients propagated to the encoder. We empirically demonstrate that the optimal $\gamma$ varies between datasets and detectors due to the number of epochs required for each one, whereas if the model requires more training epochs, the $\gamma$ should be higher.

\begin{table}[!htp]
    \centering
    \caption{Comparison with different multimodal works on RGB/IR benchmarks.}
    \resizebox{1.0\columnwidth}{!}{%
    \begin{tabular}{@{}lcccccccc@{}}
        
        \toprule

        \multirow{2}{*}[-1em]{\textbf{Method}}  &  \multicolumn{4}{c}{$\qquad$$\qquad$$\qquad$$\qquad$\textbf{Dataset}} \\
        \cmidrule(lr){2-7}
        \addlinespace[5pt]
        
        {} & \multicolumn{3}{c}{\multirow{2}{*}[1em]{\textbf{FLIR}}} & \multicolumn{3}{c}{\multirow{2}{*}[1em]{\textbf{LLVIP}}} \\
        
        \cmidrule(lr){2-4}
        \cmidrule(lr){5-7}
        
       {} &   \textbf{AP$_{50}$} & \textbf{AP$_{75}$} & \textbf{AP} & \textbf{AP$_{50}$} & \textbf{AP$_{75}$} & \textbf{AP} \\
                
        \midrule

        Halfway F.~\cite{zhang2020multispectral} & 71.5  & 31.1 & 35.8 & 91.4 & 60.1 & 55.1  \\

        GAFF~\cite{9423251} & 74.6 & 31.3 & 37.4 & 94.0 & 60.2 & 55.8 \\
        
        ProbEn~\cite{chen2022multimodal} & 75.5 & 31.8 & 37.9 & 93.4 & 50.2 & 51.5 \\

        CSSA~\cite{10209020}  & 79.2 & 37.4 & 41.3 & 94.3 & 66.6 & 59.2 \\
        
        CFT~\cite{qingyun2021cross}  & 78.7 & 35.5 & 40.2 & 97.5 & 72.9 & 63.6 \\

        DIVFusion~\cite{tang2023divfusion}  & - & - & - & 89.8 & - & 52.0 \\

        RSDet~\cite{zhao2024removal}  & 81.1 & - & 41.4 & 95.8 & - & 61.3 \\

        CrossFormer~\cite{lee2024crossformer}  & 79.3 & 38.5 & 42.1 & 97.4 & 75.4 & 65.1 \\

        \midrule
     
        \rowcolor[HTML]{EFEFEF} 
        \textbf{MiPa (Ours)} & \textbf{81.3}  &  \textbf{41.8} & \textbf{44.8}  & \textbf{98.2}  & \textbf{78.1} & \textbf{66.5} \\
    
        \bottomrule
    \end{tabular}
}
\label{tab:sota}
\end{table}

\subsection{Comparison with different RGB/IR Competitors.}

In this section, we compare our approach in terms of detection performance with other strong methods in the literature that use RGB/IR modalities. ~\tref{tab:sota} shows that MiPa is a competitive method under RGB/IR benchmarks. For instance, on FLIR, MiPa has $81.3$ AP$_{50}$, while CSSA~\cite{10209020} has $79.2$, ProbEn~\cite{chen2022multimodal} has $75.5$, GAFF~\cite{9423251} $74.6$ and Halfway Fusion~\cite{zhang2020multispectral} $71.5$, RSDet~\cite{zhao2024removal} $81.1$ and CrossFormer~\cite{lee2024crossformer} $79.3$. Furthermore, we report competitive results on LLVIP, which can be seen as the IR people detection performance over different benchmarks inclusively; both modalities are used during training and inference, which is not our case (as we just use the IR modality in this comparison for inference).

\section{Conclusion}
\label{sec:conclusion}

In this work, we have introduced a novel training method leveraging a patch-based strategy using a single vision encoder for OD to consolidate the mutual information between different modalities. This method, named MiPa, has enabled two different object detectors, DINO~\cite{zhang2022dino} and Deformable DETR~\cite{zhu2020deformable}, to achieve \textit{modality invariance} on LLVIP and FLIR datasets without having to make any specific changes for each modality, for example, additional parameters for each modality, to their architecture or increase the testing inference time. Additionally, MiPa outperformed competitors on the LLVIP and FLIR datasets. Furthermore, we provide a definition from information theory regarding the knowledge captured by the MiPa method. In future works, we plan on exploring strategies for the initial pre-train, such as MultiMAE~\cite{bachmann2022multimae}, and additionally understand at which points we can apply curriculum learning for balancing the modalities while exploring the complementary information of the modalities.

\section{Supplementary Material: {\color{red}Mi}xed {\color{red}Pa}tch Visible-Infrared Modality Agnostic Object Detection}

In this supplementary material, we provide additional information to reproduce our work. The source code is provided alongside the supplementary material, and we are going to provide the official repository. This supplementary material is divided into the following sections: Detailed diagrams (Section~\ref{sec:detailed_diagrams}), Towards the optimal \(\rho\) (Section~\ref{sec:towards}), Ablation on $\gamma$ (Section~\ref{ref:ablationongamma}) and MiPa on different detectors (Section~\ref{ref:mipavsbackbones}).

\section{Detailed diagrams}
\label{sec:detailed_diagrams}

\begin{figure*}[!h]
    \centering
    
    \includegraphics[width=0.8\textwidth]{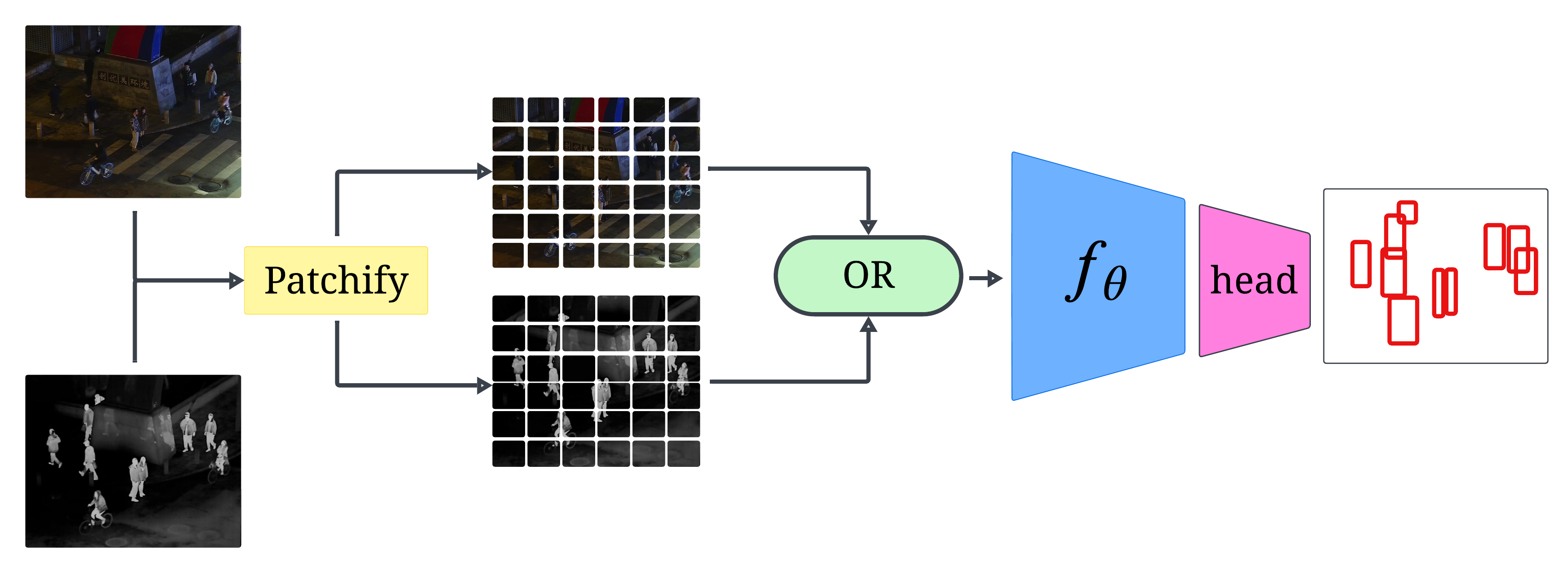}

    \caption{Our \textit{Both} baseline for multimodal object detection learning with patches. The yellow block is the patchify function. In green, we have the block representing one or the other patch modality to use. In blue is the backbone, and in pink is the head of the detector.}
    \label{fig:baselines}
\end{figure*}

In this section, we provide additional diagrams aimed at enhancing the comprehension of both the baselines and our method in more detail. In~\fref{fig:baselines}, we show the simple strategy for constructing a multimodal model utilizing patches; this is our \textit{Both} model in the main manuscript. First, the framework divides the images from both modalities (RGB and IR) into patches (yellow block). Subsequently, the extracted patches are fed into the backbone of the model (depicted in blue) and the head in pink.

\begin{figure*}[!h]
    \centering
    
    \includegraphics[width=0.8\textwidth]{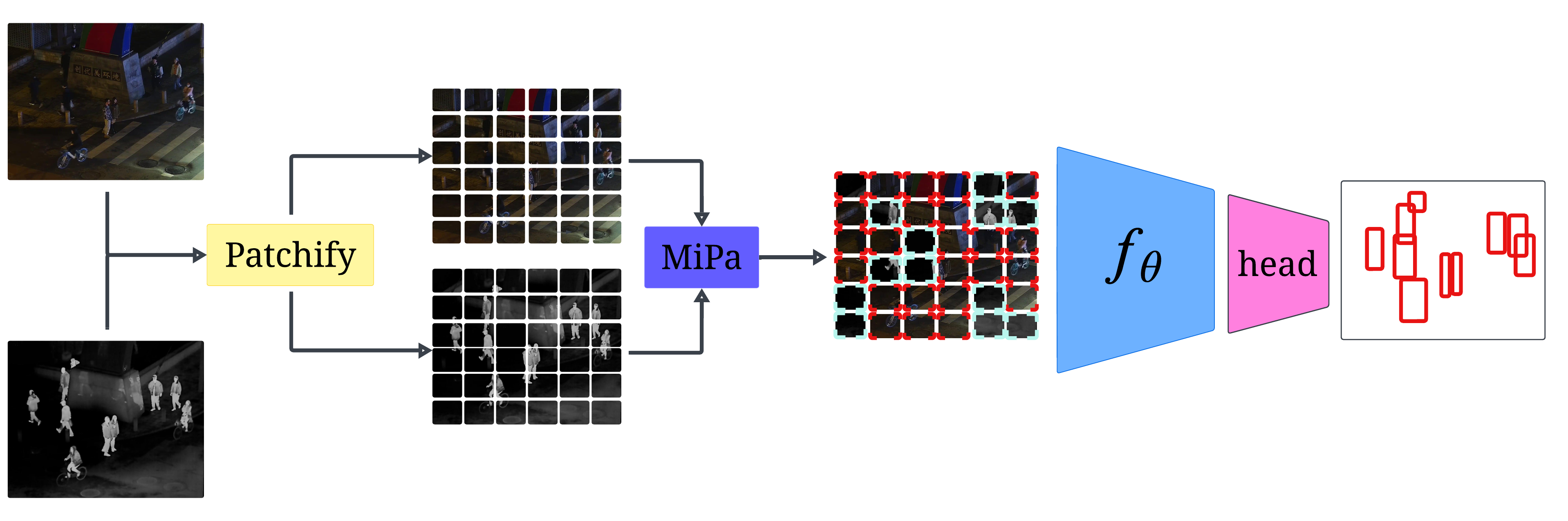}

    \caption{Mix Patches diagram: First, in yellow, is the patchify function, which is responsible for providing the patches. Second, in purple, is the mix patches function, which is responsible for mixing the patches based on a pre-defined policy, e.g., uniform distribution of both modalities. Then, in blue is the backbone, and in pink is the detection head.}
    \label{fig:mixpatches}
\end{figure*}

In~\fref{fig:mixpatches}, we present the proposed mix patches diagram. Similar to the previous diagram, we initially apply the patchify function (in yellow), followed by the mix patches function (in purple). This function receives the patches and performs a mix patches operation, such as sampling the patches from both modalities according to a uniform distribution. Finally, the backbone is illustrated in blue, and the head in pink.

\begin{figure*}[!h]
    \centering
    
    \includegraphics[width=1.0\textwidth]{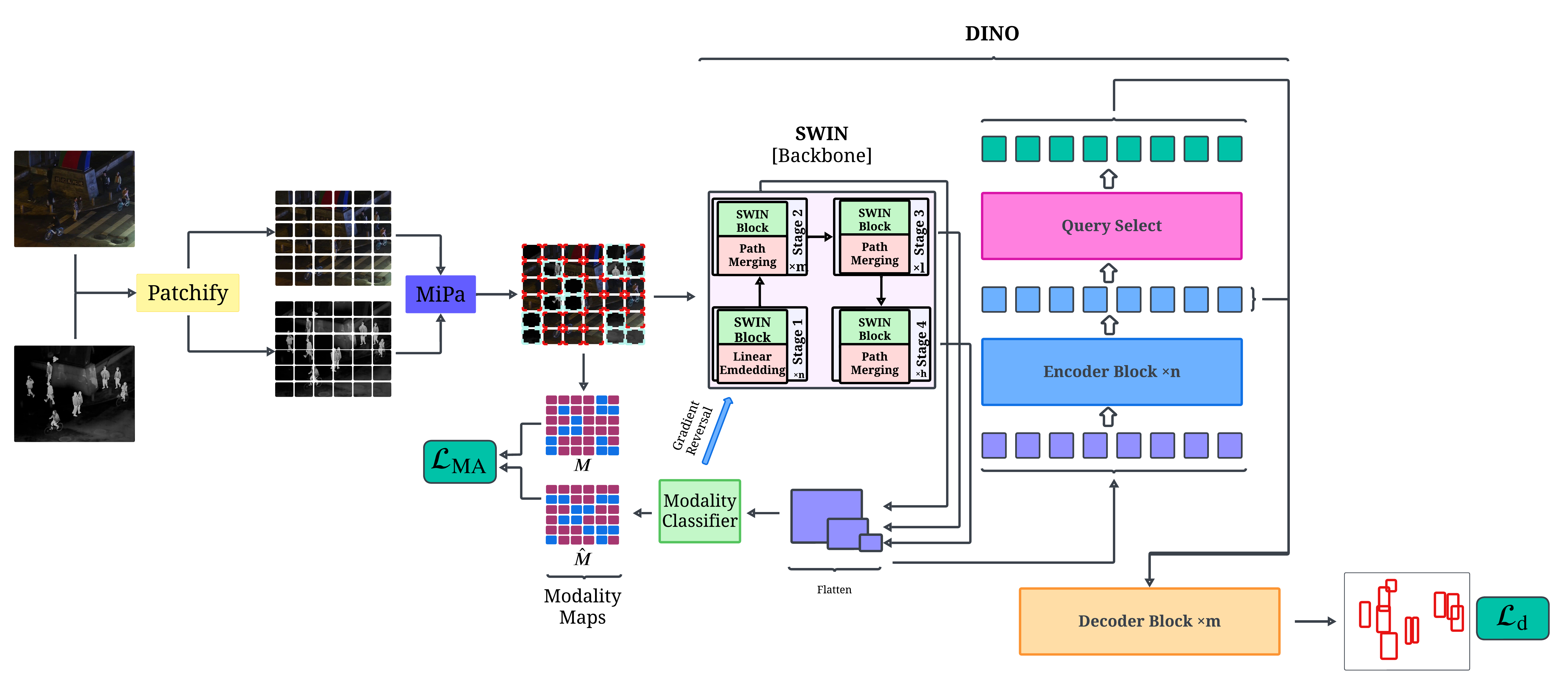}

    \caption{MiPa with DINO: First, in yellow, is the patchify function, which is responsible for providing the patches. Second, bold purple is the mixing patches function, which is responsible for mixing the patches based on a pre-defined policy, e.g., uniform distribution of both modalities. Then, we have the DINO alongside the modality classifier head for the GRL.}
    \label{fig:mipa_dino}
\end{figure*}

Lastly, we provide an overview of an implementation of MiPa with DINO in~\fref{fig:mipa_dino}. While the image is similar to the previous one, we offer additional visualizations showcasing the SWIN backbone alongside the modality classifier. For the sake of simplicity and to emphasize the MiPa's modality classifier and the patchify/mix patches components, we omit the detection head in the figure. \\

\section{Towards the optimal $\rho$}
\label{sec:towards}

In this section, similar to the main manuscript, we provide the study of various strategies devised within this work to find the optimal approach to select the parameter $\rho$. Such a parameter represents the proportion of one modality, IR in our context, sampled during the training to facilitate optimal learning. As shown in~\tref{tab:all_rho_sup}, the variable strategy yields the most favorable results in terms of providing the optimal $\rho$. This effectiveness is attributed to the inherent characteristics of MiPa to act as a regularizer for the weaker modality, which is the RGB in our setup. Thus, as described, the variable strategy is the method that reached the best average across all the different APs. For example, the variable strategy was able to reach $88.5$ AP$_{50}$ in RGB, outperforming other strategies. Although its performance in IR was slightly lower than that of the Fixed strategy [$\rho=0.25$] (achieving $97.5$ AP$_{50}$), the variable strategy's overall mean performance was superior with $93.00$ AP$_{50}$. This trend is similar to the other AP metrics, in which the RGB was improved, and the mean performance was better with the variable strategy.

\begin{table*}[!htp]
    
    \caption{Comparison of different ratio  \(\rho\) sampling methods on LLVIP. Using DINO with SWIN backbone.}
    \centering
    \resizebox{0.9\textwidth}{!}{%
    \begin{tabular}{gggggggggg}
    \toprule
        \rowcolor{white}
        {} & \multicolumn{9}{c}{\textbf{Dataset: LLVIP}} \\
        \cmidrule(lr){2-10}
        \rowcolor{white}
        \multirow{1}{*}[-0.5em]{\textbf{Model}} & \multicolumn{3}{c}{\textbf{AP$_{50}$}} & \multicolumn{3}{c}{\textbf{AP$_{75}$}} & \multicolumn{3}{c}{\textbf{AP}} \\
        
        \cmidrule(lr){2-4}
        \cmidrule(lr){5-7}
        \cmidrule(lr){8-10}
        \addlinespace[5pt]

         \cellcolor[HTML]{FFFFFF}{} &  \cellcolor[HTML]{FFFFFF}\multirow{2}{*}[1em]{\textbf{RGB}} & \cellcolor[HTML]{FFFFFF}\multirow{2}{*}[1em]{\textbf{IR}} & \cellcolor[HTML]{FFFFFF}\multirow{2}{*}[1em]{\textbf{AVG.}}  &  \cellcolor[HTML]{FFFFFF}\multirow{2}{*}[1em]{\textbf{RGB}} & \cellcolor[HTML]{FFFFFF}\multirow{2}{*}[1em]{\textbf{IR}} & \cellcolor[HTML]{FFFFFF}\multirow{2}{*}[1em]{\textbf{AVG.}}   &  \cellcolor[HTML]{FFFFFF}\multirow{2}{*}[1em]{\textbf{RGB}} & \cellcolor[HTML]{FFFFFF}\multirow{2}{*}[1em]{\textbf{IR}} & \cellcolor[HTML]{FFFFFF}\multirow{2}{*}[1em]{\textbf{AVG.}}   \\

        \midrule
       \rowcolor{white}

        Fixed [\(\rho\)=0.25]                                  & 78.9 & \textbf{98.2} & 88.55 & 41.5 & 78.1 & 59.80 & 42.5 & 66.5 & 54.50 \\
     
        Fixed [\(\rho\)=0.50] & 73.0 & 97.6 & 85.30 & 31.1 & 78.1 & 54.60 & 36.0 & 67.0 & 51.50 \\
        
        \rowcolor{white}
        Fixed [\(\rho\)=0.75] & 77.4 & 97.5 & 87.45 & 40.5 & 76.5 & 58.50 & 42.0 & 65.2 & 53.60 \\

        \midrule
        Curriculum (\(\rho\)=0.25 for 4 epochs; then variable) & 76.6 & 97.8 & 87.20 & 38.0 & 77.0 & 57.50 & 40.7 & 65.7 & 53.20 \\
        
        \rowcolor{white}
        Curriculum (\(\rho\)=0.25 for 8 epochs; then variable) & 80.1 & 97.8 & 88.95 & 40.9 & \textbf{79.1} & 60.00 & 43.0 & \textbf{67.6} & 55.30 \\

        \midrule
        Variable & \textbf{88.5} & 97.5 & \textbf{93.00} & \textbf{48.9} & 77.4 & \textbf{63.15} & \textbf{48.9} & 66.6 & \textbf{57.75} \\

    \bottomrule
   
    \end{tabular}
    }

\label{tab:all_rho_sup}
\end{table*}

\section{Ablation on $\gamma$}
\label{ref:ablationongamma}

In this section, we expand our comparison for different $\gamma$, in which we provide the full study on different AP metrics. The parameter $\gamma$ governs the rate at which the modality invariance loss influences training. Thus, for FLIR, the best $\gamma$ value was $0.05$. As shown in the~\tref{tab:ablationongamma}, we study various values of $\gamma$ with steps of $0.05$, selected following the GRL equation described in our manuscript and inspired by previous works~\cite{ganin2015unsupervised}. In this study, the values vary between $0.05$ and $0.40$, but the values may vary depending on the necessary number of epochs for training, as this function is step-dependent during training. Models that require more epochs may have larger values for $\gamma$. On FLIR, MiPa [$\gamma=0.05$] was able to outperform the other baselines with an average of $67.62$ AP$_{50}$, which is an increase from normal MiPa with $66.52$ and the best baseline with $64.72$ (Both [$\rho=0.50$]). Moreover, MiPa [$\gamma=0.05$] reached $29.77$ in terms of AP$_{75}$, which is an average increase from $29.25$ of normal MiPa, and $27.45$ from the best baseline (Both [$\rho=0.75$]). Note that for such a case, Both [$\rho=0.75$] was better in terms of localization (AP$_{75}$) in comparison with Both [$\rho=0.50$], even though it is worse than normal MiPa and MiPa with modality agnostic layer. Finally, in terms of AP, the trend is similar, so on average, we outperform all baselines and normal MiPa, which means that we are better in terms of localization and classification in each modality simultaneously. Thus, in this section, our goal of reaching a better balance between modalities while creating a robust model is successfully achieved. \\

\begin{table*}[!h]
    \caption{Comparison of detection performance over different baselines and MiPa for DINO with SWIN. The evaluation is done for RGB, IR, and the average of the modalities.}
    \centering
    \resizebox{1.0\textwidth}{!}{%
    \begin{tabular}{lcqgggggggggggggg}

        \toprule
        \rowcolor{white}
        \multirow{4}{*}[-0.5em]{\textbf{Model}} & \multirow{4}{*}[-0.5em]{\textbf{Backbone}} & {} & {} & \multicolumn{8}{c}{\textbf{Test Set (Dataset: FLIR)}} \\
        
        \cmidrule(lr){4-13}
        \addlinespace[5pt]
        
        {} & {} & \cellcolor[HTML]{FFFFFF} \multirow{1}{*}[+0.0em]{\textbf{Modality}} &  \multicolumn{3}{c}{\multirow{2}{*}[1em]{\textbf{RGB}}} & \multicolumn{3}{c}{\multirow{2}{*}[1em]{\textbf{IR}}} & \multicolumn{4}{c}{\multirow{2}{*}[1em]{\textbf{Average}}}   \\
        \cmidrule(lr){4-6}
        \cmidrule(lr){7-9}
        \cmidrule(lr){10-13}
        \addlinespace[5pt]

        \rowcolor{white}
        {} &  {} & {} &  \textbf{AP$_{50}\uparrow$} & \textbf{AP$_{75}\uparrow$} & \textbf{AP${}\uparrow$} & \textbf{AP$_{50}\uparrow$} & \textbf{AP$_{75}\uparrow$} & \textbf{AP${}\uparrow$} & \textbf{AP$_{50}\uparrow$} & \textbf{AP$_{75}\uparrow$} & \textbf{AP${}\uparrow$} \\

        \midrule

       \rowcolor{white}
        {} & {} & RGB & 66.07 ± 0.98 &	27.97 ± 0.22 &	32.33 ± 0.47 &	56.60 ± 0.80 &	20.87 ± 0.56 &	26.30 ± 0.19 &	61.33 &	24.42 &	29.32 \\

        {} & {} & IR & 56.47 ± 0.79 &	17.00 ± 0.98 &	24.30 ± 0.69 &	70.40 ± 0.38 &	38.80 ± 0.66 &	38.97 ± 0.31 &	63.43 &	27.90 &	31.63 \\
        
        \rowcolor{white}
        \multirow{5}{*}[-1.6em]{\textbf{DINO}} & \multirow{5}{*}[-1.6em]{\textbf{SWIN}}  & Both [$\rho=0.25$]  &  56.53 ± 0.76 &	18.33 ± 0.55 &	25.60 ± 0.33 &	67.57 ± 1.73 &	31.33 ± 2.10 &	34.87 ± 1.35 &	62.05 &	24.83 &	30.23 \\		
        
        {} & {} & Both [$\rho=0.50$]  & 60.50 ± 0.66 &	19.60 ± 1.29 &	27.37 ± 0.58 &	68.93 ± 0.60 &	33.03 ± 1.32 &	35.90 ± 0.82 &	64.72 &	26.32 &	31.63 \\	
        
        \rowcolor{white}
        {} & {} & Both [$\rho=0.75$]   & 58.53 ± 0.92 &	19.40 ± 0.83 &	26.47 ± 0.75 &	70.43 ± 0.65 &	35.50 ± 1.23 &	37.53 ± 0.41 &	64.48 &	27.45 &	32.00 \\	
        
        \cmidrule(lr){3-13}
        {} & {} & MiPa & 63.53 ± 1.94 &	22.33 ± 0.82 &	29.47 ± 0.92 &	69.50 ± 1.84 &	36.17 ± 0.46 &	37.57 ± 0.67 &	66.52 &	29.25 &	33.52 \\		
		
        \rowcolor{white}
        {} & {} & MiPa [$\gamma=0.05$] & 64.80 ± 2.30 &	24.77 ± 1.05 &	30.60 ± 0.62 &	70.43 ± 0.53 &	34.77 ± 1.18 &	37.50 ± 0.43 &	\textbf{67.62} &	\textbf{29.77} &	\textbf{34.05} \\															
        
        {} & {} & MiPa [$\gamma=0.10$] & 64.03 ± 2.11 &	24.10 ± 1.63 &	30.63 ± 1.22 &	69.63 ± 1.45 &	33.13 ± 1.95 &	36.80 ± 1.39 &	66.83 &	28.62 &	33.72 \\																																	
        
        \rowcolor{white}
        {} & {} & MiPa [$\gamma=0.15$] & 64.27 ± 0.47 &	24.40 ± 0.93 &	30.07 ± 0.68 &	69.93 ± 1.02 &	33.83 ± 1.24 &	36.80 ± 0.86 &	67.10 &	29.12 &	33.43 \\																																	
        {} & {} & MiPa [$\gamma=0.20$] & 61.83 ± 1.39 &	22.83 ± 1.01 &	28.53 ± 0.76 &	69.27 ± 1.57 &	31.87 ± 2.02 &	35.73 ± 1.31 &	65.55 &	27.35 &	32.13 \\																	

        \rowcolor{white}
        {} & {} & MiPa [$\gamma=0.30$] & 62.20 ± 2.49 &	22.93 ± 1.35 &	29.10 ± 1.28 &	67.47 ± 2.04 &	32.53 ± 0.66 &	35.87 ± 0.69 &	64.83 &	27.73 &	32.48 \\															

        {} & {} & MiPa [$\gamma=0.40$] & 61.13 ± 2.88 &	22.30 ± 0.57 &	28.50 ± 0.99 &	67.93 ± 0.92 &	32.47 ± 0.48 &	35.87 ± 0.49 &	64.53 &	27.38 &	32.18 \\

    \bottomrule

    \end{tabular}
    }
\label{tab:ablationongamma}
\end{table*}

\section{MiPa on different detectors}
\label{ref:mipavsbackbones}

In this section, we present additional quantitative results, including various performance metrics measured in terms of different APs. In~\tref{tab:swin}, we outline the results obtained using the SWIN backbone for DINO and Deformable DETR across baselines, MiPa, and MiPa with a modality invariance layer. As shown, MiPa demonstrates superior performance compared to using both modalities jointly and other baselines across different datasets.

\begin{table*}[!h]
    
    \caption{Comparison of detection performance over different baselines and MiPa for DINO and Deformable DETR. The evaluation is done for RGB, IR, and the average of the modalities.}
    \centering
    \resizebox{1.0\textwidth}{!}{%
    \begin{tabular}{lcqgggggggggggggg}

            \midrule
       \rowcolor{white}
        \multirow{4}{*}[-0.5em]{\textbf{Model}} & \multirow{4}{*}[-0.5em]{\textbf{Backbone}} & {} & {} & \multicolumn{8}{c}{\textbf{Dataset: LLVIP}} \\
        
        \cmidrule(lr){4-13}
        \addlinespace[5pt]
        
        {} & {} & \cellcolor[HTML]{FFFFFF} \multirow{1}{*}[+0.0em]{\textbf{Modality}} &  \multicolumn{3}{c}{\multirow{2}{*}[1em]{\textbf{RGB}}} & \multicolumn{3}{c}{\multirow{2}{*}[1em]{\textbf{IR}}} & \multicolumn{4}{c}{\multirow{2}{*}[1em]{\textbf{Average}}}   \\
        \cmidrule(lr){4-6}
        \cmidrule(lr){7-9}
        \cmidrule(lr){10-13}
        \addlinespace[5pt]

        \rowcolor{white}
        {} &  {} & {} &  \textbf{AP$_{50}\uparrow$} & \textbf{AP$_{75}\uparrow$} & \textbf{AP${}\uparrow$} & \textbf{AP$_{50}\uparrow$} & \textbf{AP$_{75}\uparrow$} & \textbf{AP${}\uparrow$} & \textbf{AP$_{50}\uparrow$} & \textbf{AP$_{75}\uparrow$} & \textbf{AP${}\uparrow$} \\

        \midrule
        
       \rowcolor{white}
        \multirow{5}{*}[-1.2em]{\textbf{DINO}} & \multirow{5}{*}[-1.2em]{\textbf{SWIN}} & RGB & 90.87 ± 0.84 &	54.20 ± 1.02 &	51.87 ± 0.79 &	94.23 ± 0.57 &	67.13 ± 0.85 &	59.43 ± 0.48 &	92.55 &	60.67 &	55.65 \\																									
        {} & {} & IR & 66.87 ± 0.90 &	20.27 ± 0.98 &	29.03 ± 0.76 &	96.87 ± 0.12 &	73.53 ± 0.40 &	64.27 ± 0.12 &	81.87 &	46.90 &	46.65 \\														
        \rowcolor{white}
        {} & {} & Both [$\rho=0.25$] & 79.73 ± 1.03 &	45.70 ± 0.43 &	44.97 ± 0.33 &	97.40 ± 0.22 &	76.03 ± 0.83 &	65.87 ± 0.45 &	88.57 &	60.87 &	55.42 \\										
        {} & {} & Both [$\rho=0.50$] & 82.40 ± 1.50 &	47.27 ± 1.65 &	46.43 ± 1.03 &	96.50 ± 0.29 &	74.17 ± 2.10 &	64.83 ± 0.96 &	89.45 &	60.72 &	55.63 \\		
        
        \rowcolor{white}
        {} & {} & Both [$\rho=0.75$]   & 81.23 ± 2.89 &	45.60 ± 2.49 &	45.23 ± 2.13 &	97.07 ± 0.25 &	74.73 ± 1.41 &	65.27 ± 0.82 &	89.15 &	60.17 &	55.25 \\		
        
        \cmidrule(lr){3-13}
        {} & {} & \textbf{MiPa (Ours)} &  88.70 ± 0.45 &	46.67 ± 0.86 &	48.00 ± 0.28 &	96.97 ± 0.26 &	73.07 ± 1.42 &	64.30 ± 1.10 & \textbf{92.83} &	59.87 &	\textbf{56.15} \\

        {} & {} & \textbf{MiPa + MI (Ours)} & 89.10 ± 0.28 &	46.60 ± 0.86 &	48.10 ± 0.33 &	96.83 ± 0.09 &	71.17 ± 0.70 &	63.17 ± 0.58 &	\textbf{92.97} &	58.88 &	55.63 \\																
        \midrule

          \rowcolor{white}
        \multirow{5}{*}[-1.2em]{\textbf{Def.DETR}} & \multirow{5}{*}[-1.2em]{\textbf{SWIN}} & RGB  & 80.00 ± 1.50 &	35.50 ± 0.22 &	40.27 ± 0.41 &	90.03 ± 0.87 &	50.37 ± 0.85 &	49.67 ± 0.48 &	85.02 &	42.93 &	\textbf{44.97} \\																							

        {} & {} & IR & 56.10 ± 2.50 &	10.77 ± 1.47 &	21.10 ± 1.34 &	94.20 ± 0.08 &	62.20 ± 0.86 &	56.73 ± 0.47 &	75.15 &	36.48 &	38.92 \\																																
        
        \rowcolor{white}
        {} & {} & Both [$\rho=0.25$]  & 51.20 ± 3.47 &	22.57 ± 1.96 &	25.70 ± 1.91 &	83.73 ± 16.57 &	54.17 ± 16.62 &	48.30 ± 14.93 &	67.47 &	38.37 &	37.00 \\																													
        
        {} & {} & Both [$\rho=0.50$]   & 53.57 ± 4.17 &	23.13 ± 2.15 &	26.57 ± 2.11 &	83.87 ± 16.17 &	52.67 ± 17.17 &	49.37 ± 12.64 &	68.72 &	37.90 &	37.97 \\																														
        
        \rowcolor{white}
        {} & {} & Both [$\rho=0.75$]  & 53.53 ± 4.55 &	22.83 ± 2.72 &	26.5 ± 2.63 &	82.33 ± 18.48 &	51.33 ± 18.56 &	48.13 ± 14.03 &	67.93 &	37.08 &	37.32 \\																													
        
        \cmidrule(lr){3-13}
        {} & {} & \textbf{MiPa (Ours)} &  78.60 ± 0.42 &	23.33 ± 5.85 &	29.20 ± 6.37 &	95.20 ± 0.16 &	62.60 ± 0.78 &	56.80 ± 0.45 &	\textbf{86.90} &	\textbf{42.97} &	43.00 \\																		

        {} & {} & \textbf{MiPa + MI (Ours)} &  79.02 ± 0.21 &	24.36 ± 2.85 &	31.25 ± 4.32 &	95.36 ± 0.25 &	63.38 ± 0.43 &	57.25 ± 0.43 &	\textbf{87.19} &	\textbf{43.87} &	44.25 \\

        \toprule
        \rowcolor{white}
        \multirow{4}{*}[-0.5em]{\textbf{Model}} & \multirow{4}{*}[-0.5em]{\textbf{Backbone}} & {} & {} & \multicolumn{8}{c}{\textbf{Dataset: FLIR}} \\
        
        \cmidrule(lr){4-13}
        \addlinespace[5pt]
        
        {} & {} & \cellcolor[HTML]{FFFFFF} \multirow{1}{*}[+0.0em]{\textbf{Modality}} &  \multicolumn{3}{c}{\multirow{2}{*}[1em]{\textbf{RGB}}} & \multicolumn{3}{c}{\multirow{2}{*}[1em]{\textbf{IR}}} & \multicolumn{4}{c}{\multirow{2}{*}[1em]{\textbf{Average}}}   \\
        \cmidrule(lr){4-6}
        \cmidrule(lr){7-9}
        \cmidrule(lr){10-13}
        \addlinespace[5pt]

        \rowcolor{white}
        {} &  {} & {} &  \textbf{AP$_{50}\uparrow$} & \textbf{AP$_{75}\uparrow$} & \textbf{AP${}\uparrow$} & \textbf{AP$_{50}\uparrow$} & \textbf{AP$_{75}\uparrow$} & \textbf{AP${}\uparrow$} & \textbf{AP$_{50}\uparrow$} & \textbf{AP$_{75}\uparrow$} & \textbf{AP${}\uparrow$} \\

        \midrule

       \rowcolor{white}
        \multirow{5}{*}[-1.2em]{\textbf{DINO}} & \multirow{5}{*}[-1.2em]{\textbf{SWIN}} & RGB & 66.07 ± 0.98 &	27.97 ± 0.22 &	32.33 ± 0.47 &	56.60 ± 0.80 &	20.87 ± 0.56 &	26.30 ± 0.19 &	61.33 &	24.42 &	29.32 \\

        {} & {} & IR & 56.47 ± 0.79 &	17.00 ± 0.98 &	24.30 ± 0.69 &	70.40 ± 0.38 &	38.80 ± 0.66 &	38.97 ± 0.31 &	63.43 &	27.90 &	31.63 \\
        
        \rowcolor{white}
        {} & {} & Both [$\rho=0.25$]  &  56.53 ± 0.76 &	18.33 ± 0.55 &	25.60 ± 0.33 &	67.57 ± 1.73 &	31.33 ± 2.10 &	34.87 ± 1.35 &	62.05 &	24.83 &	30.23 \\		
        
        {} & {} & Both [$\rho=0.50$]  & 60.50 ± 0.66 &	19.60 ± 1.29 &	27.37 ± 0.58 &	68.93 ± 0.60 &	33.03 ± 1.32 &	35.90 ± 0.82 &	64.72 &	26.32 &	31.63 \\	
        
        \rowcolor{white}
        {} & {} & Both [$\rho=0.75$]   & 58.53 ± 0.92 &	19.40 ± 0.83 &	26.47 ± 0.75 &	70.43 ± 0.65 &	35.50 ± 1.23 &	37.53 ± 0.41 &	64.48 &	27.45 &	32.00 \\	
        
        \cmidrule(lr){3-13}
        {} & {} & \textbf{MiPa (Ours)} & 63.53 ± 1.94 &	22.33 ± 0.82 &	29.47 ± 0.92 &	69.50 ± 1.84 &	36.17 ± 0.46 &	37.57 ± 0.67 &	\textbf{66.52} &	\textbf{29.25} & \textbf{33.52} \\												

        {} & {} & \textbf{MiPa + MI (Ours)} & 64.80 ± 2.30 &	24.77 ± 1.05 &	30.60 ± 0.62 &	70.43 ± 0.53 &	34.77 ± 1.18 &	37.50 ± 0.43 &	\textbf{67.62} &	\textbf{29.77} &	\textbf{34.05} \\

         \midrule
       \rowcolor{white}
        \multirow{5}{*}[-1.2em]{\textbf{Def.DETR}} & \multirow{5}{*}[-1.2em]{\textbf{SWIN}} & RGB & 49.33 ± 1.39 &	13.93 ± 0.30 &	20.97 ± 0.53 &	43.77 ± 0.56 &	10.13 ± 0.08 &	17.37 ± 0.19 &	46.55 &	12.03 &	19.17 \\																	

        {} & {} & IR & 39.17 ± 1.48 &	08.57 ± 0.24 & 14.90 ± 0.50 & 59.20 ± 0.29 &	20.03 ± 0.33 & 26.93 ± 0.62 & 49.18 &	14.30 &	20.92 \\																																	
        
        \rowcolor{white}
        {} & {} & Both [$\rho=0.25$]  & 35.73 ± 4.95 &	08.27 ± 1.51 &	14.00 ± 2.38 &	43.00 ± 13.54 &	14.30 ± 5.97 &	19.23 ± 7.01 &	39.37 &	11.28 &	16.62 \\																	
        
        {} & {} & Both [$\rho=0.50$]  & 33.93 ± 5.15 &	08.23 ± 1.43 &	13.60 ± 2.17 &	43.33 ± 14.14 &	14.70 ± 6.34 &	19.63 ± 7.43 &	38.63 &	11.47 &	16.62 \\																																	
        \rowcolor{white}
        {} & {} & Both [$\rho=0.75$]   & 32.90 ± 3.54 &	07.70 ± 1.20 &	12.97 ± 1.65 &	44.13 ± 14.85 &	14.17 ± 6.30 &	19.47 ± 7.37 &	38.52 &	10.93 &	16.22 \\																												
        \cmidrule(lr){3-13}
        
        {} & {} & \textbf{MiPa (Ours)} &  48.00 ± 0.57 &	15.23 ± 0.69 &	20.70 ± 0.45 &	54.97 ± 0.90 &	19.80 ± 0.28 &	25.50 ± 0.42 &	\textbf{51.48} &	\textbf{17.52} &	\textbf{23.10} \\
        
        {} & {} & \textbf{MiPa + MI (Ours)} &  48.27 ± 1.76 &	14.57 ± 1.05 &	20.63 ± 0.96 &	55.80 ± 0.22 &	21.00 ± 0.67 &	26.33 ± 0.39 &	\textbf{52.03} &	\textbf{17.78} &	\textbf{23.48} \\					

    \bottomrule

    \end{tabular}
    }

\label{tab:swin}
\end{table*}

{\small
\bibliographystyle{ieee_fullname}
\bibliography{egbib}
}

\end{document}